\documentclass{article}

\ifdefined\neuripscameraready
  \usepackage[main,final]{neurips_2026}
\else
  \usepackage[preprint]{neurips_2026}
  \makeatletter
  \renewcommand{\@noticestring}{Preprint. Accepted at the 40th Conference on Neural Information Processing Systems (NeurIPS 2026).}
  \makeatother
\fi

\usepackage[utf8]{inputenc}
\usepackage[T1]{fontenc}
\usepackage{courier}

\usepackage{graphicx}
\usepackage{multirow}
\usepackage{soul}
\usepackage{booktabs}
\usepackage[table]{xcolor}
\definecolor{tablerowgray}{HTML}{E6E6E6}
\usepackage{pifont}
\usepackage{marvosym}
\usepackage{float}
\usepackage{microtype}
\usepackage{url}
\usepackage{amsfonts}
\usepackage{nicefrac}

\usepackage[accsupp]{axessibility}  
\usepackage{xspace} 
\usepackage[ruled]{algorithm2e}

\usepackage{hyperref}

\IfFileExists{orcidlink.sty}{%
  \usepackage{orcidlink}%
}{%
  \newcommand{\orcidlink}[1]{}%
}

\newif\ifglossary\glossarytrue   

\newcommand{\cmark}{\ding{51}}
\newcommand{\xmark}{\ding{55}}

\begin{document}

\title{\textit{MvFFN}: Multi-view Floor plan Feed-Forward Network for Unposed Wide-Baseline Panorama Layout Reconstruction} 


\author{%
  Yuguang Li\textsuperscript{1}\thanks{Equal contribution.\quad \protect\Letter\ Corresponding author: \texttt{ylee3@cs.washington.edu}.}\hspace{0.45em}\textsuperscript{\Letter} \quad
  Yichuan Deng\textsuperscript{1}\footnotemark[1] \quad
  Zixuan Liu\textsuperscript{1}\footnotemark[1] \quad
  Ivaylo Boyadzhiev\textsuperscript{2} \\
  \textbf{Xiuyuan Wu\textsuperscript{1} \quad
  Xinyu Zou\textsuperscript{1} \quad
  Linda Shapiro\textsuperscript{1} \quad
  Alex Colburn\textsuperscript{1}} \\
  \textsuperscript{1}University of Washington \quad
  \textsuperscript{2}Independent Researcher
}


\maketitle
\begin{abstract}Reconstructing accurate camera poses and floor plan layouts from sparse, unposed, wide-baseline RGB panoramas remains a challenging open problem. Direct multi-room floor plan prediction is especially difficult in this setting: pose, geometry, semantics, and global layout structure are tightly coupled, yet a complete floor plan is hard to learn as a single global output. We address this challenge with \textit{Multi-view Ordered Wall Instance Segmentation} (\textit{MOIS}), a new formulation that replaces direct global wall-sequence prediction with local cross-view wall instances and within-room wall connectivity.
Building on this formulation, we propose \textit{MvFFN}, a feed-forward multi-view transformer that jointly predicts coarse camera poses, dense global geometry, and wall-instance-level semantic outputs with globally consistent identities and within-room connectivity. A downstream \textit{Layout Chaining} (\textit{LC}) post-processing pipeline then assembles these predictions into multi-room floor plans with shared global coordinates, consistent wall identities, and connected room structure. The resulting pipeline substantially outperforms prior methods on wall- and junction-level accuracy. Moreover, the predicted wall instances enable fully automatic downstream layout refinement, with planar bundle adjustment improving a majority of scenes and yielding clear gains on the subset that benefits. We also introduce \texttt{ZInD-CrossView}, a re-annotated dataset with globally unique wall instances and cross-view segmentation labels for this task.
\end{abstract}

\section{Introduction}
\label{sec:intro}

Reconstructing floor plan layouts and camera poses from panoramic images is important for applications such as augmented reality, indoor digital twins, immersive walkthrough generation, and autonomous navigation. Recovering complete multi-room layouts requires multiple views because of occlusions and limited visibility. However, existing methods typically avoid the setting of unposed, wide-baseline panoramas: they assume known camera poses~\cite{yue2023roomformer, avetisyan2024scenescript, liu2024polyroom}, require small-baseline captures~\cite{avetisyan2024scenescript}, or rely on human-in-the-loop adjustment~\cite{cruz2021zillow, li2025badgr}. A fully automatic approach for reconstructing layouts and poses from unposed, wide-baseline panoramas would substantially reduce capture and annotation effort and enable broader deployment at scale.

A floor plan encodes wall geometry, local connectivity, and room-level organization. Recovering it from multiple views requires associating wall instances across views, inferring their connectivity despite occlusion, and separating structures that belong to different rooms. Geometry alone is often insufficient. When camera poses are unknown and fused point maps are noisy, reconstruction often suffers from duplicate walls, missing walls, and incorrect junctions, which reduce both wall and junction-level precision and recall. In contrast, images provide critical semantic cues for grouping pixels across views by wall instance, inferring local connectivity, and resolving room assignments. At the same time, directly predicting a complete multi-room floor plan as a single global output is difficult to learn. Large multi-room layouts often contain ambiguous room partitions, limited observations can obscure small structural details, and small local sequence errors can cascade into large global drift. The problem is further compounded by the scarcity of datasets that provide structured multi-room layouts together with camera-registered multi-view panoramas. This motivates the central insight of our paper: rather than predicting a global floor plan directly, we recast multi-view, multi-room layout estimation around local wall relations that are easier to supervise and more robust to noisy annotations.

To this end, we study structured floor plan and camera-pose reconstruction from sparse, unposed 360$^\circ$ panoramas, producing complete layouts when view coverage is sufficient and partial layouts under occlusion. We formalize this idea as \textit{Multi-view Ordered Wall Instance Segmentation} (\textit{MOIS}), which replaces direct global floor plan prediction with local wall instances and within-room wall connectivity that remain consistent across views. Based on \textit{MOIS}, we propose \textit{MvFFN}, a feed-forward multi-view network that predicts pixel-grounded wall instances and their within-room connectivity from cross-view geometric and semantic cues, and a \textit{Layout Chaining} (\textit{LC}) post-processing pipeline that assembles these local predictions into multi-room floor plans with consistent wall identities in a shared global frame. \textit{MvFFN} also predicts coarse camera poses and dense global geometry, while the inferred wall instances enable downstream layout bundle adjustment for further refinement. Together, these design choices make multi-view layout estimation more learnable under scarce data and noisy supervision.

The contributions of this work are as follows:
(1) We introduce \textit{MOIS}, a new formulation for multi-view, multi-room layout estimation based on local cross-view wall instances and within-room wall connectivity.
(2) We introduce \textit{MvFFN}, a feed-forward multi-view network that predicts these local wall structures from cross-view geometric and semantic cues. We show gains both before and after layout assembly: dense \textit{MOIS} predictions improve raw cross-view wall-slot quality, while the assembled global layouts substantially improve wall and junction accuracy over prior modular baselines across both 5-view and 8-view settings.
(3) We introduce \texttt{ZInD-CrossView} (\texttt{ZInD-XV}) dataset with globally unique wall instances and cross-view segmentation annotations.

\section{Related Work}
\textbf{Layout estimation.} Most layout estimation methods combine a dense geometric front-end with a back-end that vectorizes predictions into wall and corner instances and room topology. In the single-view setting, methods predict monocular 1D floor boundaries~\cite{yang2019dula, wang2021led2, jiang2022lgtnet, sun2021hohonet} and recover wall and corner sequences through post-processing~\cite{zou2018layoutnet, sun2019horizonnet, pintore2020atlantanet}, relying on the clockwise image-column order within a single panorama. This cue does not directly extend to sparse multi-view settings, where each view observes only a partial layout under occlusion. For multi-room floor-plan estimation, most methods instead assume accurate camera poses and dense point clouds, and infer structure mainly from fused geometric cues in top-down space. Under imperfect geometry, connectivity is typically recovered with heuristic optimization~\cite{chen2019floorsp, wang2022psmnet} or learned polygon and graph decoders~\cite{liu2024polyroom, yue2023roomformer, chen2023_polydiffuse, su2023slibo}. By contrast, cross-view semantic cues remain underused for resolving wall identity and connectivity under pose uncertainty. \textit{SceneScript}~\cite{avetisyan2024scenescript} shows that, given camera poses and RGB video, semantic context can improve structured layout estimation.

\noindent\textbf{Wide-baseline pose and geometry estimation.} Wide-baseline indoor pose estimation remains difficult for traditional \textit{Structure-from-Motion}~\cite{schonberger2016colmap, moulon2016openmvg, lambert2022salve}. Recent learning-based models improve dense 3D point map and camera pose estimation from wide baseline inputs using strong foundation model encoders~\cite{wang2024dust3r, wang2025vggt, tang2025mvdust3r}. For indoor $360^\circ$ panoramas, limited training data often constrains geometry estimation to 1D floor boundary representations rather than dense per-pixel 3D geometry~\cite{hutchcroft2022covispose, tu2024panopose, nejatishahidin2023graph}. \textit{BADGR}~\cite{li2025badgr} shows that semantic layout correspondences and geometry are complementary: once pixel-to-wall correspondences and initial estimates are available from unposed wide-baseline images, poses and layouts can be jointly optimized to reduce initial errors.

\section{Overview}
\begin{figure}[t]
\begin{center}
    \includegraphics[width=1.0\linewidth]{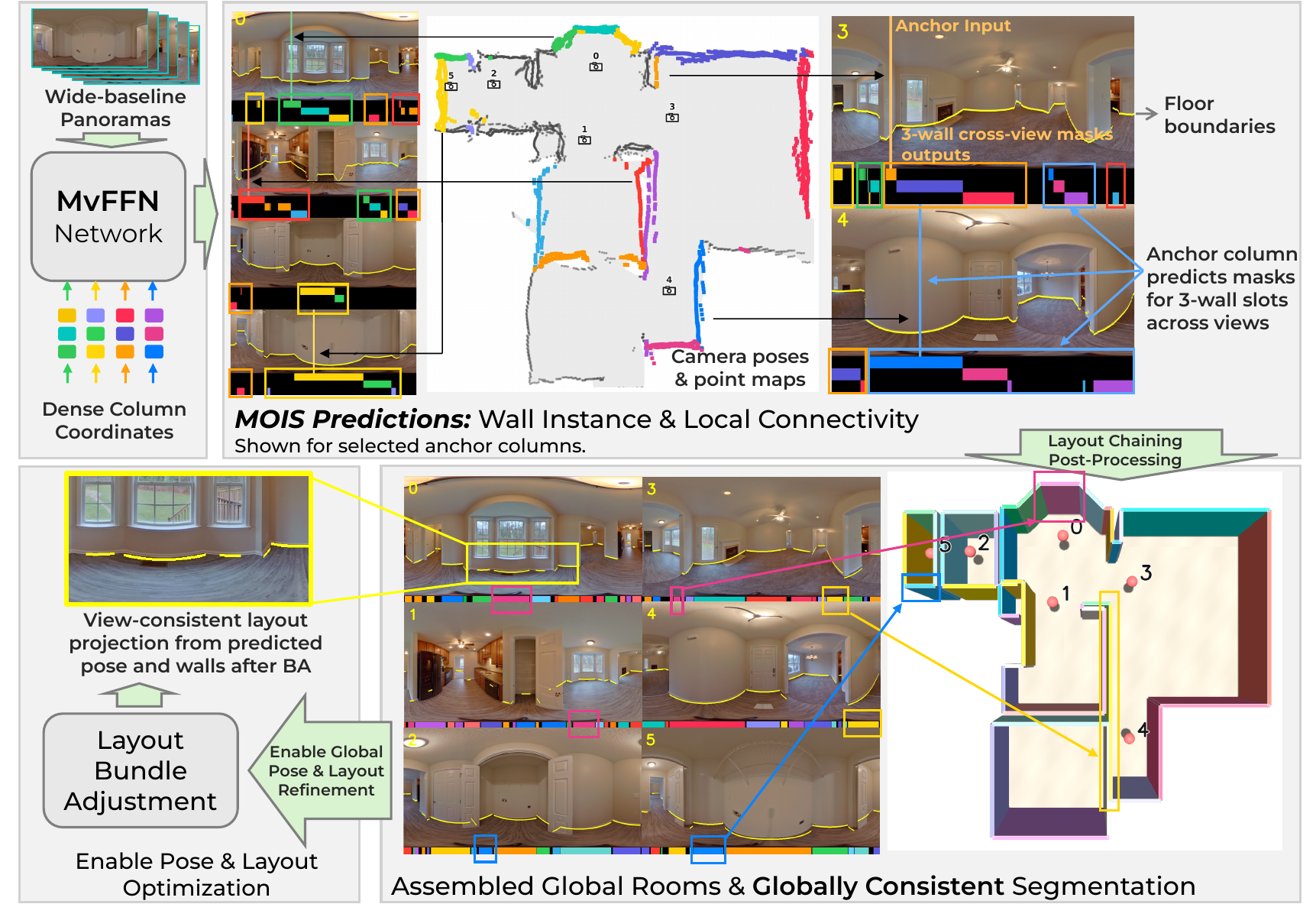}
    \vspace{-20pt}
    \caption{
    \textit{MvFFN} reformulates floor plan reconstruction from direct global prediction to local \textit{MOIS} prediction: each anchor column, shown as a vertical line in the panorama, predicts cross-view masks for the anchor wall and two clockwise adjacent walls. Colors indicate corresponding wall slots/instances across panoramas and in the assembled layout. \textit{Layout Chaining} aggregates these local predictions into global wall identities, room-level wall sequences, and metric floor plan layouts, enabling downstream refinement.
    }
    \label{fig:overview}
    \vspace{-16pt}
\end{center}
\end{figure}

Large multi-room floor plans often contain ambiguous room boundaries, occluded wall connectivity, and distant structural details, making direct floor-level sequence prediction brittle to error accumulation. We therefore reconstruct layouts through dense local \textit{MOIS} predictions rather than direct global floor plan prediction.



Our bottom-up pipeline (Figure~\ref{fig:overview}) takes \(N\) panoramas and evaluates \textit{MvFFN} densely at every column of every view. Each anchor column defines one \textit{MOIS} target: the cross-view assignments of the current wall and its within-room neighbors. Aggregating all anchors yields dense anchor-conditioned \textit{MOIS} assignments over multiple views, predicted jointly with global geometry under pose uncertainty.


\textit{MvFFN} uses a multi-view geometric encoder, trained from scratch and inspired by \textit{VGGT}~\cite{wang2025vggt}, adapted to panoramic 2D column tokens and dense column-wise outputs. A floor-layout semantic branch combines image and geometric features for local wall-instance reasoning. We describe the geometric encoder and floor-layout decoder in Sec.~\ref{sec:encoder} and Sec.~\ref{sec:decoder}, respectively.


Finally, the dense local \textit{MOIS} predictions are assembled into global room layouts via \textit{Layout Chaining} (Sec.~\ref{sec:postproc}). The predicted pixel-to-wall correspondences and initial layouts can also support downstream layout refinement, which we evaluate with layout bundle adjustment in Sec.~\ref{sec:ba}.

\section{Method}
We formulate multi-room layout reconstruction from sparse panoramas~\cite{zhang2014panocontext} as the recovery of global 3-DoF camera poses (2D position and 1D rotation), structured 2D floor layouts, and a local anchor-conditioned wall representation across views.

\noindent\textbf{Inputs.} We assume a set of \(N\) straightened \(360^\circ\) RGB panoramas \(\{I_i\}_{i=1}^N\)~\cite{zhang2014panocontext}, where each \(I_i \in \mathbb{R}^{3 \times H \times W}\) is given in equirectangular projection. The panoramas are sparsely captured in an indoor environment and are assumed to form a single pose graph whose neighboring views have at least $20\%$ co-visibility similar to~\cite{cruz2021zillow, hutchcroft2022covispose, li2025badgr}. We further assume that relative camera heights are known across the floor.

\noindent\textbf{\textit{MvFFN} outputs.} For each panorama \(I_i\), \textit{MvFFN} predicts four intermediate quantities: (1) a 3-DoF camera pose \(E_i \in SE(2)\); (2) dense per-column floor boundaries \(\mathcal{B}_i\in\mathbb{R}^{2 \times W}\); (3) dense per-column floor point maps \(\mathcal{P}_i\in\mathbb{R}^{4 \times W}\) and wall-line parameters \(\Lambda_i\in\mathbb{R}^{3 \times W}\); and (4) for each anchor column, a dense \textit{MOIS} probability field \(\hat{\mathcal{S}}\) over columns across all views, encoding assignments to the underlying global wall instance intersected by the anchor ray and its within-room neighboring wall slots.

\label{sec:mois}
\noindent\textbf{\textit{MOIS} formulation.} 
We formalize the local wall-instance and within-room connectivity predictions described above with \textit{MOIS}, an anchor-conditioned ordered wall representation defined for each image column (see Fig.~\ref{fig:overview} for intuition). Let \(\mathcal{C}_{i,c}\) denote anchor column \(c\) in panorama \(I_i\). The anchor ray determines the current room by the first wall it intersects: if it intersects a closed door, the door is assigned to the attached wall instance and the current room remains the attached room; if it passes through an open doorway, the current room is the room containing the first visible wall it intersects. The corresponding \textit{MOIS} target is then a local ordered \(K\)-wall tuple
\[
\mathbf{w}_{i,c} = \left(w_{i,c}^{(0)}, w_{i,c}^{(1)}, \dots, w_{i,c}^{(K-1)}\right),
\]
where \(w_{i,c}^{(0)}\) is the intersected wall instance defining the current room, and \(w_{i,c}^{(r)}\) denotes the \(r\)-th clockwise physically adjacent wall within that room. We convert this local tuple into a dense binary assignment tensor \(\mathcal{S}^{\mathrm{gt}}\), where each entry indicates whether column \((j,d)\) is assigned to a given ordered wall slot of the anchor column. This dense local representation serves as the semantic target for the floor-layout semantic branch in Sec.~\ref{sec:decoder} and is assembled into global layouts by \textit{Layout Chaining} in Sec.~\ref{sec:postproc}.

\noindent\textbf{Final layout outputs.} The primary outputs are 3-DoF camera poses \(\{E_i\}_{i=1}^N\) and global room layouts \(\{\mathcal{W}^m\}\). Unlike the local anchor-conditioned \textit{MOIS} tuples, each \(\mathcal{W}^m\) represents the final reconstructed wall geometry of room \(m\) in the global floor coordinate frame after \textit{Layout Chaining}. We also maintain an auxiliary column-to-wall map \(\mathcal{M}:\mathcal{C}_{i,c}\rightarrow w^\ell\), which associates each image column with at most one global wall instance. The room corners \(V^m=(v^{m,1},\dots,v^{m,K_m})\) are obtained by intersecting consecutive wall segments, where \(v^{m,k}\in\mathbb{R}^2\) and $K_m$ is the number of corners. Depending on visibility, a room layout may be closed or remain open under occlusion.

\subsection{N-view Geometry Encoder Branch}
\label{sec:encoder}

The geometry branch is a multi-view encoder for dense geometric prediction, including camera pose, floor-boundary depth, and point maps. Architecturally, it is similar to \textit{VGGT}~\cite{wang2025vggt}, while operating on panoramic column-wise features. Each panorama $I_i$ is processed by an \textit{EfficientNet-v2} backbone~\cite{tan2021efficientnetv2}, followed by a Height Compression Module (HCM)~\cite{sun2019horizonnet} that produces column-patch features $f_i^G \in \mathbb{R}^{S \times C}$, where $S=256$, $C=1024$, and each patch spans $4 \times H$ pixels. 

\noindent\textbf{Patch encodings.} We add angular column embeddings (ACE) and sinusoidal camera positional encodings (CPE)~\cite{vaswani2017attention} to obtain geometry tokens $F^G$. The ACE are \textit{Fourier features}~\cite{tancik2020fourier} of per-patch viewing angles. For each column-patch, the viewing angle combines: (1) an intra-frame horizontal angle $\theta^h_{i,c} \in [0^\circ, 360^\circ)$ corresponding to the patch-center azimuth, and (2) a precomputed vanishing angle $\theta_i^{v} \in [-45^\circ, 45^\circ)$ for camera $i$~\cite{zhang2014panocontext}. Incorporating vanishing-angle embeddings exploits the high precision of indoor vanishing angles~\cite{li2025badgr, lambert2022salve} to axis-align the predicted floor layout. The global floor coordinate frame is centered at the first camera and oriented by its vanishing angle.
 
\noindent\textbf{Alternating-attention transformer.} We concatenate geometry tokens from all views, $t^G = \bigcup_{i=1}^N {f_i^G}$, with camera tokens $t_i^C \in \mathbb{R}^{1 \times C}$ and register tokens $t_i^R \in \mathbb{R}^{4 \times C'}$~\cite{darcet2023registertokens}, and process them with a 12-layer alternating-attention transformer~\cite{wang2025vggt}. The transformer alternates between frame-wise and global attention layers implemented with \textit{FlashAttention}~\cite{dao2022flashattention}, and produces updated geometry, camera, and register tokens $(\hat{t}_i^G, \hat{t}_i^C, \hat{t}_i^R)_{i=1}^N$.

\noindent\textbf{Dense predictions.} We use separate multi-layer perceptron (MLP) heads to predict dense floor boundaries \(\hat{\mathcal{B}}\), 2D floor point maps \(\hat{\mathcal{P}}\), and 2D wall-line parameters \(\hat{\Lambda} \). The point-map output contains both 2D floor coordinates and their 2D aleatoric uncertainties. Each supporting line is represented by a 2D orientation vector and a bias term. Floor boundaries and point maps are predicted with aleatoric uncertainty~\cite{kendall2016aleatoricuncertainty}. Following prior panorama layout work~\cite{sun2019horizonnet,hutchcroft2022covispose,li2025badgr}, floor boundaries and point maps are supervised using derived column-wise targets that pass through open doorways and ignore occlusions from furniture and other non-structural objects, subject to dataset-specific annotation conventions. Camera poses are predicted from per-camera register and camera tokens~\cite{wang2025vggt} using an MLP head with adaptive LayerNorm~\cite{perez2018film}.

\begin{figure}[t]
\begin{center}
    \includegraphics[width=1.0\linewidth]{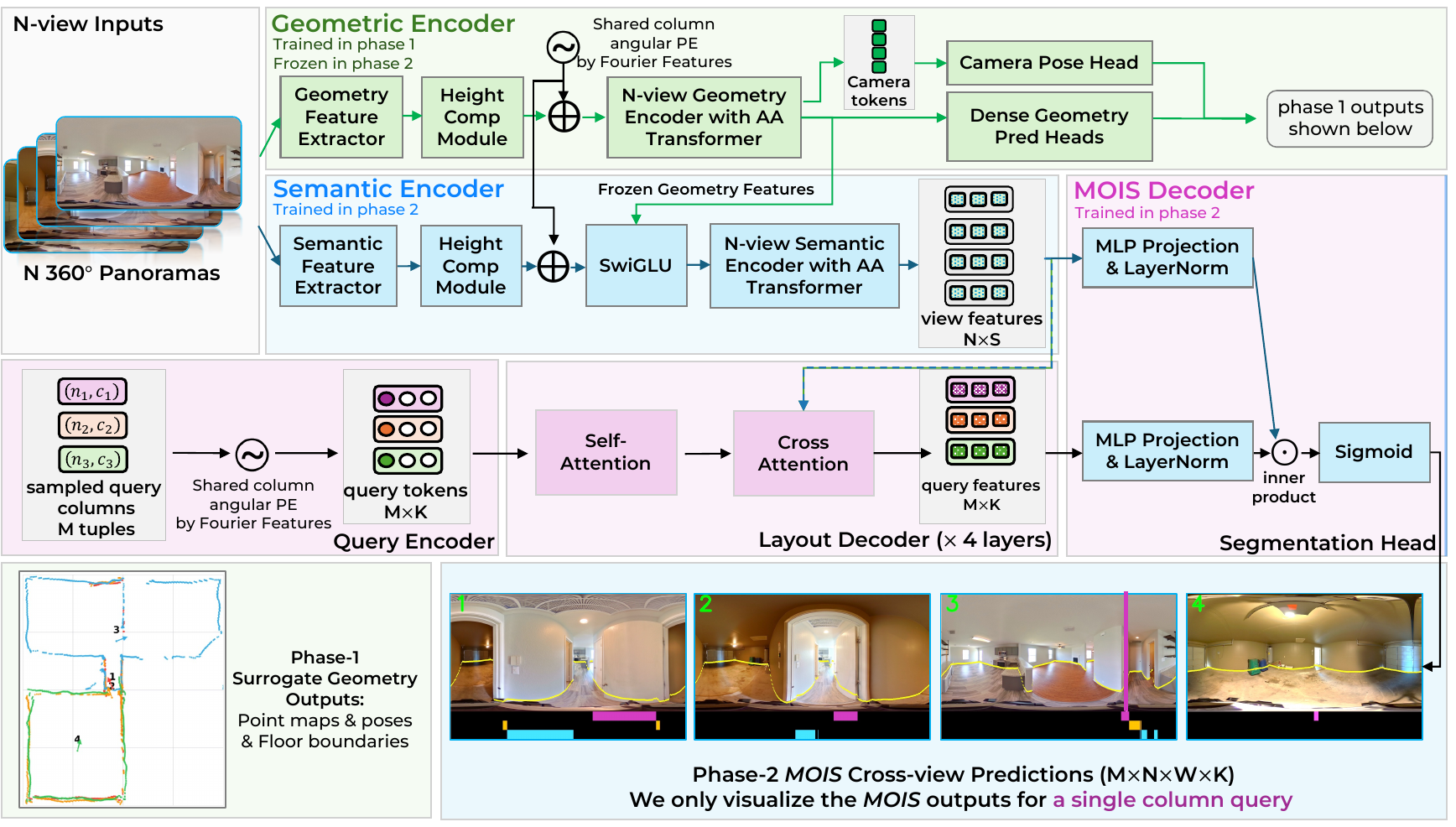}
    \vspace{-16pt}
    \caption{Architecture of \textit{MvFFN} with two-stage training for \textit{MOIS} prediction. 
    } 
    \label{fig:architecture}
    \vspace{-14pt}
\end{center}
\end{figure}

\subsection{Floor Layout Semantic Branch}
\label{sec:decoder}

Our floor-layout semantic branch predicts the dense \textit{MOIS} probability field introduced in Sec.~\ref{sec:mois}. Specifically, for each anchor column, it estimates which columns across all views belong to the current wall and to its next $K-1$ clockwise physically adjacent within-room neighbors. 
We use $K=3$, which provides enough neighboring-wall context for stable connectivity recovery in \textit{Layout Chaining} (Sec.~\ref{sec:postproc}), while avoiding the degradation of longer prediction windows (see Appendix~\ref{sec:mois_window_ablation}).
The branch consists of a semantic $N$-view encoder conditioned on geometry features, a cross-attention transformer decoder with anchor-conditioned wall-slot queries, and a segmentation head.

\paragraph{Semantic \(N\)-view encoder.}
The semantic encoder follows the geometry branch architecture from Sec.~\ref{sec:encoder}. Each panorama is processed by an \textit{EfficientNet-v2} backbone and HCM to produce column-patch features \(f_i^S \in \mathbb{R}^{S \times C}\). These features are fused with the frozen geometry tokens \(\hat{t}_i^G\) through a SwiGLU module, and the fused features are then processed by a four-layer semantic alternating-attention transformer, producing semantic column tokens \(\hat{t}_i^S\).

\paragraph{Anchor-conditioned wall-slot queries.}
We adopt a two-level query design with \(M \times K\) learned query tokens, where \(M\) is the number of sampled anchor columns and \(K\) is the number of ordered \textit{MOIS} wall slots. Each query group corresponds to one anchor column and contains \(K\) ordered query tokens associated with the wall slots \(r \in \{0,\dots,K-1\}\). Given sampled anchor columns \(\{(i_m,c_m)\}_{m=1}^M\), we initialize query-group features using the angular encoding of the anchor column together with a camera identity encoding.

\paragraph{Transformer decoder.}
The decoder has four stacked layers. Within each anchor group \(m\), self-attention operates over the \(K\) ordered wall-slot queries, and cross-attention is applied to the semantic column tokens from all views. The decoder outputs updated query features \(\hat{t}_{m,r}^q\), one for each anchor \(m\) and wall slot \(r\).

\paragraph{\textit{MOIS} probability prediction.}
Let \(\tilde{t}_{m,r}^q \in \mathbb{R}^{C'}\) and \(\tilde{t}_{j,d}^I \in \mathbb{R}^{C'}\) denote the projected query and semantic features for anchor-slot \((m,r)\) and column \((j,d)\), respectively. We compute the dense \textit{MOIS} probability field as
\begin{equation}
\hat{\mathcal{S}}_{m,j,d,r}
=
\sigma\!\left(\left\langle \tilde{t}_{m,r}^q,\tilde{t}_{j,d}^I \right\rangle\right),
\end{equation}
where \(\hat{\mathcal{S}}_{m,j,d,r}\) is the probability that column \((j,d)\) belongs to the \(r\)-th ordered wall slot of anchor \(m\). During training, \(M\) anchor columns are randomly sampled subject to GPU memory limits. During inference, we set \(M=NW\) and predict the dense \textit{MOIS} field for all anchor columns in all views in a single forward pass; equivalently a 5D tensor indexed by view and anchor column, as in Algorithm~\ref{alg:layout_chaining}.

\subsection{Training}
\label{sec:training}


We train \textit{MvFFN} in two stages because the geometry and semantic objectives compete under joint training. In the first stage, we train the geometry branch for dense geometric prediction, including floor boundaries, floor point maps, and camera poses, using a multi-task objective that combines geometric supervision with absolute and relative pose losses. In the second stage, we freeze the pre-trained geometry branch and train the floor-layout semantic branch from scratch under dense \textit{MOIS} supervision, reusing the geometry features from stage one. The \textit{MOIS} objective is factorized into a wall-presence term, which predicts whether a column belongs to any ordered wall slot of the anchor, and an ordered wall-slot classification term applied to positive assignments. In both stages, training samples are formed by drawing sparse multi-view panorama groups from a single pose graph with sufficient co-visibility. Full loss definitions, optimizer settings, and sampling schedules are provided in Appendix~\ref{sec:training_details}.

\subsection{Inductive Bias for Training Data Scarcity}

\label{sec:inductive_bias}

Training sparse multi-view models is substantially harder than training 2-view models because they require more diverse multi-view samples, while the number of scenes and unique structured training examples remains limited. In our settings, this makes naive 3-view to 5-view training unstable even when the same data is sufficient for strong 2-view geometric learning. This trend is also consistent with recent geometric models such as \textit{VGGT}~\cite{wang2025vggt} and \textit{MV-DUSt3R}~\cite{tang2025mvdust3r}, which require substantially more training data than \textit{DUSt3R}~\cite{wang2024dust3r} to improve 2-view pose accuracy. To improve convergence under this regime, we introduce two inductive biases based on reliable 2-view \textit{CovisPose}~\cite{hutchcroft2022covispose} predictions. 

\noindent\textbf{Camera rotation priors.} Two-view geometric models already provide reliable camera rotations, and per-panorama vanishing points further improve indoor rotation accuracy~\cite{zhang2014panocontext, lambert2022salve}. We first run the 2-view \textit{CovisPose} model and recover global camera rotations with a minimum spanning tree, following \textit{BADGR}'s initialization strategy~\cite{li2025badgr}. 
These refined rotations are then injected as conditioning signals into the angular encoder (Sec.~\ref{sec:encoder}). 
This reduces the amount of scarce multi-view supervision spent on relearning camera rotation and lets the model focus on the harder cross-view geometric reasoning problem, while retaining near 2-view rotation accuracy without requiring substantially more \(N\)-view training data.

\noindent\textbf{Column-patch correspondence priors.} Under limited multi-view training data, we use precomputed 2-view \textit{CovisPose} correspondences during both training and inference to restrict early cross-view attention to a local window around predicted matches, rather than relearning global correspondence search from scratch. We implement this by masking the first four global self-attention layers of the alternating-attention transformer: for each column-patch token in view $I_a$, tokens in view $I_b$ more than 40 columns away from the predicted match are masked out, and fully non-covisible view pairs are blocked entirely. This reduces correspondence ambiguity, leading to less noisy gradients and faster convergence.


\subsection{Layout Chaining}
\label{sec:postproc}


\textit{Layout Chaining} (\textit{LC}) is a deterministic procedure that reduces dense local \textit{MOIS} predictions into global column-to-wall assignments \(\mathcal{M}\), room-level wall sequences \(\{\mathcal{L}^m\}\), and room endpoints \(\{V^m\}\). Since many anchor columns provide overlapping same-wall and neighboring-wall evidence, \textit{LC} first consolidates wall identity and then recovers within-room adjacency. It clusters same-view current-wall predictions \(\hat{\mathcal{S}}[n,:,n,:,0]\) into per-view wall instances \(L_n\), then merges these across views into global wall groups through a cross-view same-wall graph. Directed within-room adjacency is estimated from the neighboring \textit{MOIS} slots, producing a room graph whose edges are filtered by geometric consistency and whose longest directed paths define partial room layouts \(\mathcal{L}^m\). Finally, wall-line parameters are fit from supporting point-map predictions grouped by \(\mathcal{M}\), and room endpoints are recovered by intersecting consecutive walls. Algorithm details are provided in Appendix~\ref{sec:lc_alg}.


\subsection{Layout Bundle Adjustment}
\label{sec:ba}

The layouts produced by \textit{Layout Chaining} and the predicted camera poses may still violate multi-view consistency and structural layout constraints. The predicted pixel-to-wall correspondences provide the associations needed for refinement. We investigate whether two refinement methods, planar bundle adjustment (\textit{BA}) and \textit{BADGR}~\cite{li2025badgr}, can use these correspondences to improve poses and wall geometry. Sec.~\ref{sec:ba_evals} reports the results.


\section{Experiments}
We evaluate primarily on \texttt{ZInD-XV}, which extends \texttt{ZInD}~\cite{cruz2021zillow} with globally indexed multi-room layouts represented as room-wise wall sequences, and per-panorama column-wise floor-boundary and wall-instance labels derived from those layouts. \texttt{ZInD-XV} contains 2,706 floors and 66,834 panoramas. We additionally use \texttt{HM3D-XV}, a smaller auxiliary benchmark derived from residential \texttt{HM3DSem} scenes~\cite{yadav2023hm3dsem}, with 372 floors and 17,943 panoramas, to evaluate cross-dataset transfer to furnished, mesh-rendered indoor scenes. Dataset construction and annotation details are provided in the supplementary material.

We compare against modular baselines assembled from existing pose, geometry, and layout components. The main evaluation measures final structured floor-layout accuracy after \textit{Layout Chaining}; we also report pose estimation and downstream refinement with planar bundle adjustment and \textit{BADGR}~\cite{li2025badgr}. To isolate the learned representation before post-processing, we additionally evaluate raw \textit{MOIS} prediction quality in the supplementary material. Baseline implementation details, \texttt{HM3D-XV} transfer results, and additional ablations are also provided in the supplementary material.

\subsection{Pose Estimation Evaluation on ZInD-XV}
\label{sec:pose_evals}

We compare \textit{MvFFN} with \textit{CovisPose}+\textit{Greedy Spanning Tree} (\textit{GST})~\cite{nejatishahidin2023graph,li2025badgr} and \textit{VGGT}~\cite{wang2025vggt} on \texttt{ZInD-XV}, adapting \textit{VGGT} to panoramas via perspective decomposition. We exclude \textit{GraphCovis}~\cite{nejatishahidin2023graph} and \textit{PanoPose}~\cite{tu2024panopose} because public code is unavailable. For each view-count setting, we evaluate on 1,000 \(N\)-view test tuples using the pose-graph alignment protocol from prior work~\cite{nejatishahidin2023graph}; tuple construction and implementation details are provided in the supplementary. Although pose estimation is not the primary target of \textit{MvFFN}, Table~\ref{tab:pose_zind_58} shows that it is comparable to \textit{CovisPose}+\textit{GST}: \textit{CovisPose}+\textit{GST} attains lower median errors, while \textit{MvFFN} reduces mean error, variance, and hard-case failures. \textit{VGGT} performs substantially worse in this sparse wide-baseline panorama setting.

\begin{table}[h]
\centering
\vspace{-4pt}
\caption{Camera pose evaluation on \texttt{ZInD-XV} dataset.}
\label{tab:pose_zind_58}
\vspace{-8pt}
\setlength{\tabcolsep}{3pt}
\renewcommand{\arraystretch}{1.12}
\resizebox{0.8\linewidth}{!}{
\begin{tabular}{c|cccccc|cccccc}
\hline
\multirow{2}{*}{Methods}        & \multicolumn{3}{c}{5v Pos (cm)$\downarrow$}  & \multicolumn{3}{c|}{5v Rot ($^\circ$)$\downarrow$} & \multicolumn{3}{c}{8v Pos (cm)$\downarrow$}   & \multicolumn{3}{c}{8v Rot ($^\circ$)$\downarrow$} \\ \cline{2-13} 
                                & Mn            & Std           & Med          & Mn              & Std            & Med            & Mn            & Std           & Med           & Mn             & Std            & Med            \\ \hline
\textit{CovisPose}+\textit{GST} & 12.6          & 23.3          & \textbf{9.3} & 1.3             & 0.9            & \textbf{0.7}   & 17.3          & 37.3          & \textbf{11.8} & 1.9            & 1.6            & \textbf{0.8}   \\
\textit{VGGT}                   & 30.7          & 39.4          & 13.3         & 11.1            & 16.0           & 4.0            & 53.4          & 63.1          & 27.8          & 13.3           & 17.7           & 4.9            \\
\textit{\textbf{MvFFN}}         & \textbf{12.5} & \textbf{10.8} & 10.2         & \textbf{1.2}    & \textbf{0.6}   & 1.1            & \textbf{17.0} & \textbf{15.6} & 13.8          & \textbf{1.3}   & \textbf{0.8}   & 1.2            \\ \hline
\end{tabular}
}
\vspace{-8pt}
\end{table}

\subsection{Floor Layout Estimation Evaluation on ZInD-XV}

\label{sec:layout_evals}

We compare \textit{MvFFN} with modular floor-layout pipelines because no prior method directly predicts structured multi-room layouts from sparse, unposed \(N\)-view panoramas. As summarized in Table~\ref{tab:floor_layout_58}, each baseline composes pose estimation, geometric input construction, and layout vectorization: \textit{CovisPose}+\textit{GST} (\textit{CP}) or ground-truth poses provide the pose input, \textit{LGT-Net} (\textit{LGT}) provides panorama-level boundary cues, and the final layout is produced by either heuristic \textit{LGT} post-processing (\textit{LGT-PP}) or learning-based top-down vectorization with \textit{RoomFormer} (\textit{RF})~\cite{yue2023roomformer} and \textit{PolyRoom} (\textit{PR})~\cite{liu2024polyroom}. We exclude \textit{VGGT} from layout evaluation because its pose accuracy is substantially worse than \textit{CovisPose}; implementation and metric details are provided in the supplementary material. 

\begin{table}[h]
\centering
\vspace{-4pt}
\caption{Floor-layout evaluation on \texttt{ZInD-XV}. Distance is in cm. Geom. denotes the intermediate geometric input and Layout denotes the vectorization backend. GT-pose denotes a control group.}
\label{tab:floor_layout_58}
\setlength{\tabcolsep}{3pt}
\renewcommand{\arraystretch}{1.12}
\resizebox{\linewidth}{!}{
\begin{tabular}{ccc|cccc|cccc|cccc|cccc}
\hline
\multicolumn{3}{c|}{Modular Pipeline}                                                  & \multicolumn{4}{c|}{5-view wall}                              & \multicolumn{4}{c|}{5-view junction}                          & \multicolumn{4}{c|}{8-view wall}                              & \multicolumn{4}{c}{8-view junction}                           \\
\multicolumn{1}{c|}{pose}        & \multicolumn{1}{c|}{geom.}        & layouts         & Prec.$\uparrow$          & Rec.$\uparrow$        & F1 $\uparrow$            & Dist $\downarrow$          & Prec.$\uparrow$          & Rec.$\uparrow$        & F1 $\uparrow$            & Dist $\downarrow$          & Prec.$\uparrow$          & Rec.$\uparrow$        & F1 $\uparrow$            & Dist $\downarrow$          & Prec.$\uparrow$          & Rec.$\uparrow$        & F1 $\uparrow$            & Dist $\downarrow$          \\ \hline
\multicolumn{3}{c|}{\textit{\textbf{MvFFN}}}                                           & \textbf{88.5} & \textbf{87.7} & \textbf{88.1} & \textbf{13.9} & \textbf{79.2} & \textbf{68.2} & \textbf{73.3} & \textbf{13.8} & \textbf{89.4} & \textbf{87.5} & \textbf{88.4} & \textbf{19.0} & \textbf{81.0} & \textbf{69.8} & \textbf{75.0} & \textbf{19.1} \\ \cline{1-3}
\multicolumn{2}{c|}{\textit{CP}}                                     & \textit{RF}     & 31.9          & 44.2          & 37.1          & 84.9          & 15.4          & 20.9          & 17.7          & 137.8         & 35.0          & 42.4          & 38.3          & 79.0          & 15.7          & 18.6          & 17.0          & 136.8         \\ \cline{1-3}
\multicolumn{2}{c|}{\textit{CP}}                                     & \textit{PR}     & 33.6          & 29.3          & 31.3          & 86.6          & 17.4          & 14.9          & 16.1          & 129.7         & 31.8          & 25.0          & 28.0          & 86.2          & 13.8          & 10.5          & 11.9          & 133.4         \\ \cline{1-3}
\multicolumn{1}{c|}{\textit{CP}} & \multicolumn{1}{c|}{\textit{LGT}} & \textit{LGT-PP} & 75.6          & 71.4          & 73.4          & 21.3          & 60.0          & 54.8          & 57.3          & 21.3          & 76.3          & 72.7          & 74.5          & 24.4          & 61.7          & 56.7          & 59.1          & 24.4          \\ \cline{1-3}
\rowcolor{tablerowgray}
\multicolumn{1}{c|}{GT}          & \multicolumn{1}{c|}{\textit{LGT}} & \textit{LGT-PP} & 78.8          & 70.5          & 74.4          & 19.2          & 62.4          & 54.3          & 58.1          & 19.7          & 80.5          & 70.8          & 75.3          & 22.7          & 65.1          & 55.5          & 59.9          & 22.7          \\ \hline
\end{tabular}
}
\end{table}

We evaluate wall and junction precision, recall, and \(F1\) with instance-level matching, and report matched-wall localization error; details are in the supplementary material. These structure-aware metrics are better suited to our setting than region \(\mathrm{IoU}\), since high-quality floor plan reconstruction requires accurate wall instances, junctions, connectivity, and geometric alignment, and \textit{MvFFN} may produce either closed polygons or open wall sequences under limited view coverage~\cite{cruz2021zillow}. As shown in Table~\ref{tab:floor_layout_58}, \textit{MvFFN} substantially outperforms all modular baselines across both 5-view and 8-view settings, with the largest gains over learned top-down vectorization backends and consistent improvements over the strongest \textit{LGT-PP} baseline. Notably, \textit{MvFFN} also outperforms \textit{LGT-PP} with ground-truth poses, indicating that the gains are not explained merely by better pose estimation. Instead, the wall and junction improvements show that cross-view \textit{MOIS} prediction and \textit{Layout Chaining} recover wall identities, within-room connectivity, and global layout structure more reliably than modular geometry-first pipelines.


\subsection{Global Refinement}


We evaluate whether \textit{MvFFN}'s predicted wall instances and pixel-to-wall correspondences support automatic downstream refinement. Table~\ref{tab:ba_zind} reports matched-wall distance before and after planar bundle adjustment (\textit{BA}) and \textit{BADGR}~\cite{li2025badgr}, both over all samples and over the subset improved by each method. Planar bundle adjustment improves a majority of samples in both 5-view and 8-view settings, with clear gains on the improved subset. Overall gains are modest, and \textit{BADGR} is more sensitive to initialization because it assumes complete initial room layouts and clean correspondences. Details are provided in the supplementary material.

\label{sec:ba_evals}
\begin{table}[h]
\centering
\vspace{-4pt}
\caption{Matched-wall distance before and after refinement on \texttt{ZInD-XV}. Distances are in cm.}
\label{tab:ba_zind}
\setlength{\tabcolsep}{3pt}
\renewcommand{\arraystretch}{1.12}
\resizebox{0.77\linewidth}{!}{
\begin{tabular}{cc|cccc|ccccc}
\hline
\multirow{3}{*}{View}  & \multirow{3}{*}{Method}      & \multicolumn{4}{c|}{All samples}                                            & \multicolumn{5}{c}{Improved samples}                                                                                                                             \\ \cline{3-11} 
                        &                               & \multicolumn{2}{c|}{Mean dist}                & \multicolumn{2}{c|}{Med. dist} & \multicolumn{1}{c|}{\multirow{2}{*}{\begin{tabular}[c]{@{}c@{}}Improved\\ (\%)\end{tabular}}} & \multicolumn{2}{c|}{Mean dist}                & \multicolumn{2}{c}{Med. dist} \\
                        &                               & before & \multicolumn{1}{c|}{after}         & before     & after            & \multicolumn{1}{c|}{}                                                                        & before & \multicolumn{1}{c|}{after}         & before     & after           \\ \hline
\multirow{2}{*}{5-view} & \textit{MvFFN}+\textit{BA}    & 13.9   & \multicolumn{1}{c|}{\textbf{13.7}} & 6.9        & \textbf{6.7}     & \multicolumn{1}{c|}{\textbf{68.0\%}}                                                         & 14.8   & \multicolumn{1}{c|}{\textbf{12.3}} & 7.5        & \textbf{5.9}    \\
                        & \textit{MvFFN}+\textit{BADGR} & 13.9   & \multicolumn{1}{c|}{17.7}          & 6.9        & 8.1              & \multicolumn{1}{c|}{\textbf{45.9\%}}                                                         & 16.2   & \multicolumn{1}{c|}{\textbf{14.1}} & 8.2        & \textbf{6.2}    \\ \hline
\multirow{2}{*}{8-view} & \textit{MvFFN}+\textit{BA}    & 19.0   & \multicolumn{1}{c|}{\textbf{18.8}} & 8.8        & 8.8              & \multicolumn{1}{c|}{\textbf{66.0\%}}                                                         & 20.6   & \multicolumn{1}{c|}{\textbf{17.8}} & 9.7        & \textbf{8.2}    \\
                        & \textit{MvFFN}+\textit{BADGR} & 19.0   & \multicolumn{1}{c|}{21.0}          & 10.1       & 10.4             & \multicolumn{1}{c|}{\textbf{48.5\%}}                                                         & 22.9   & \multicolumn{1}{c|}{\textbf{20.0}} & 8.8       & \textbf{8.8}    \\ \hline
\end{tabular}
\vspace{-52pt}
}
\end{table}

\section{Ablation Studies}



We evaluate three complementary ablations. The main paper studies \textbf{architecture}, testing whether the two-branch, two-stage \textit{MvFFN} design is necessary for stable \textit{MOIS} prediction. Appendix~\ref{sec:inductive_bias_ablation} studies \textbf{inductive bias}, evaluating rotation and correspondence priors across input-view counts, and Appendix~\ref{sec:mois_window_ablation} studies \textbf{prediction window}, where longer \textit{MOIS} wall sequences become harder to learn.

\paragraph{Architecture.}
\label{sec:ablation_architecture}
We ablate two core design choices in \textit{MvFFN}: a dedicated semantic branch for \textit{MOIS} prediction and the two-stage geometry-to-semantics training strategy. Table~\ref{tab:ablation_arch} compares four variants: a shared-encoder model, a two-branch model without geometry-to-semantic fusion, a full architecture trained in one stage, and the full two-stage \textit{MvFFN}. The results show that both geometry-conditioned semantic reasoning and staged training are critical: removing geometry-to-semantic fusion sharply reduces recall and \(F1\), while shared features or single-stage training underperform and become unstable, indicating tension between dense geometry objectives and the semantic \textit{MOIS} objective. The full two-stage design gives the best stability and final layout accuracy.

\begin{table}[H]
\centering
\vspace{-8pt}
\caption{Ablation study of \textit{MvFFN} on \texttt{ZInD-XV}, comparing encoder design, geometry fusion, and staged training. -- denotes pose metrics identical to \textit{MvFFN}.}
\label{tab:ablation_arch}
\setlength{\tabcolsep}{3pt}
\renewcommand{\arraystretch}{1.12}
\resizebox{\linewidth}{!}{
\begin{tabular}{c|cccc|ccc|ccc|cccc|cccc}
\hline
\multirow{2}{*}{Method} & \multirow{2}{*}{\begin{tabular}[c]{@{}c@{}}Geom \\ Enc\end{tabular}} & \multirow{2}{*}{\begin{tabular}[c]{@{}c@{}}Sem\\ Enc\end{tabular}} & \multirow{2}{*}{\begin{tabular}[c]{@{}c@{}}Geom\\ Fusion\end{tabular}} & \multirow{2}{*}{\begin{tabular}[c]{@{}c@{}}Train\\ Stages\end{tabular}} & \multicolumn{3}{c|}{5v pos (cm)$\downarrow$}              & \multicolumn{3}{c|}{5v rot ($^\circ$)$\downarrow$}     & \multicolumn{4}{c|}{5v wall}                                  & \multicolumn{4}{c}{5v junction}                               \\ \cline{6-19} 
                        &                                                                      &                                                                    &                                                                        &                                                                         & Mn            & Std           & Med           & Mn           & Std          & Med          & Prec$\uparrow$         & Rec$\uparrow$       & $F1\uparrow$            & Dist$\downarrow$          & Prec$\uparrow$         & Rec$\uparrow$       & $F1\uparrow$            & Dist$\downarrow$          \\ \hline
Shared-Enc              & \cmark                                                               & \xmark                                                             & \xmark                                                                 & 2                                                                       & 35.8          & 50.1          & 26.4          & 2.1          & 1.3          & 2.0          & 86.4          & 69.8          & 77.2          & 31.2          & 72.9          & 50.4          & 59.6          & 39.9          \\
No-Fusion               & \cmark                                                               & \cmark                                                             & \xmark                                                                 & 2                                                                       & --            & --            & --            & --           & --           & --           & \textbf{92.6} & 22.7          & 36.5          & 28.0          & \textbf{84.3} & 14.4          & 24.6          & 21.4          \\
1-stage                 & \cmark                                                               & \cmark                                                             & \cmark                                                                 & 1                                                                       & 46.8          & 50.5          & 30.1          & 2.5          & 1.5          & 1.9          & 87.7          & 71.9          & 79.0          & 33.9          & 75.8          & 53.1          & 62.5          & 44.1          \\
\textit{\textbf{MvFFN}} & \cmark                                                               & \cmark                                                             & \cmark                                                                 & 2                                                                       & \textbf{12.5} & \textbf{10.8} & \textbf{10.2} & \textbf{1.2} & \textbf{0.6} & \textbf{1.1} & 88.5          & \textbf{87.7} & \textbf{88.1} & \textbf{13.9} & 79.2          & \textbf{68.2} & \textbf{73.3} & \textbf{13.8} \\ \hline
\end{tabular}
\vspace{-26pt}
}
\end{table}




\section{Qualitative Results}

\label{sec:quals}


Qualitative results in Fig.~\ref{fig:quals} show representative 8-view reconstructions, with additional examples provided in Appendix~\ref{sec:supp_quals}. Across simple and complex multi-room scenes, \textit{MvFFN} produces more complete and globally consistent layouts than the strongest modular baseline, \textit{LGT-Net} with \textit{CovisPose}. The predicted wall instances are color-coded to show cross-view consistency. Compared with the baseline, \textit{MvFFN} better preserves room shape, avoids duplicate or fragmented walls, and recovers neighboring structure visible through doorways under partial occlusion.



\begin{figure}[!ht]
\begin{center}
    \vspace{-4pt}
    \includegraphics[width=1.0\linewidth]{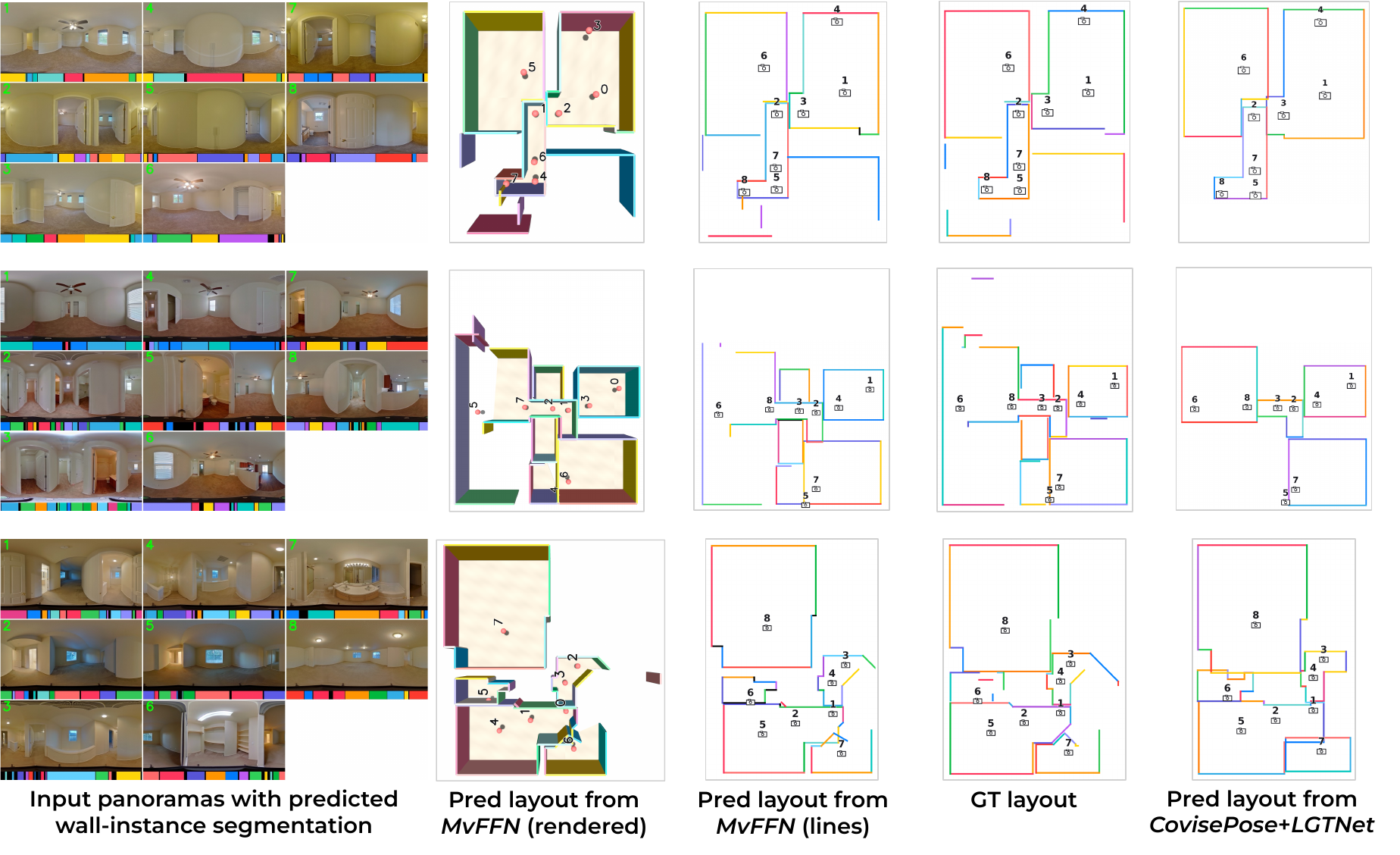}
    \vspace{-16pt}
    \caption{Qualitative comparisons on \texttt{ZInD-XV} test set. We show the strongest modular baseline, \textit{LGT-Net} with \textit{CovisPose}.} 
    \label{fig:quals}
    \vspace{-16pt}
\end{center}
\end{figure}






\section{Conclusion}

We introduced \textit{MvFFN}, which reconstructs structured floor plans and camera poses from sparse, unposed $360^\circ$ panoramas by predicting cross-view wall instances and within-room connectivity before global assembly. Experiments show clear gains over modular baselines, demonstrate automatic downstream refinement from predicted wall correspondences, and introduce \texttt{ZInD-XV} for future structured multi-view layout research.


\clearpage  


%
\bibliographystyle{plainnat}
\bibliography{main}
\clearpage
\appendix
\numberwithin{table}{section}
\numberwithin{figure}{section}
\section*{Appendix}

\section{Training Details}
\begingroup
\setlength{\abovedisplayskip}{1.5pt plus 0.5pt minus 0.5pt}
\setlength{\belowdisplayskip}{1.5pt plus 0.5pt minus 0.5pt}
\setlength{\abovedisplayshortskip}{0.5pt plus 0.5pt minus 0.25pt}
\setlength{\belowdisplayshortskip}{0.5pt plus 0.5pt minus 0.25pt}
\setlength{\parskip}{0pt}
\setlength{\jot}{1pt}

\label{sec:training_details}

We train \textit{\textit{MvFFN}} in two stages.

\paragraph{$N$-view geometry pre-training.} 
\label{sec:geometry_training}
We first train the geometry branch with the multi-task loss
\begingroup
\small
\begin{equation}
\label{equation:stage_1}
\mathcal{L}^G = \mathcal{L}_{\mathcal{P}} + \mathcal{L}_{\mathcal{B}} + \mathcal{L}_{rel} + \mathcal{L}_{abs}.
\end{equation}
\endgroup
Here, $\mathcal{L}_{\mathcal{P}}$ and $\mathcal{L}_{\mathcal{B}}$ denote the point-map and floor boundary losses. Following \textit{DUSt3R}~\cite{wang2024dust3r}, both use aleatoric uncertainty weighting~\cite{kendall2016aleatoricuncertainty} with predicted uncertainties $\Sigma_i^{\mathcal{P}}$ and $\Sigma_i^{\mathcal{B}}$. The pose losses $\mathcal{L}_{rel}$ and $\mathcal{L}_{abs}$ follow \textit{CovisPose}~\cite{hutchcroft2022covispose}; valid pairs for $\mathcal{L}_{rel}$ have minimum co-visibility 0.01. Specifically,
\begingroup
\small
\begin{equation*}
\begin{aligned}
\mathcal{L}_{\mathcal{P}} &=
\sum_{i=1}^{N} \left\lVert \Sigma_i^{\mathcal{P}} (\mathcal{P}_i - \hat{\mathcal{P}}_i) \right\rVert
- \alpha \log \Sigma_i^{\mathcal{P}}, \\
\mathcal{L}_{\mathcal{B}} &=
\sum_{i=1}^{N} \left\lVert \Sigma_i^{\mathcal{B}} (\mathcal{B}_i - \hat{\mathcal{B}}_i) \right\rVert
- \alpha \log \Sigma_i^{\mathcal{B}}, \\
\mathcal{L}_{abs} &=
\sum_{i=1}^{N} \left(\lVert t_i - \hat{t}_i \rVert + \beta_r \lVert r_i - \hat{r}_i \rVert\right).
\end{aligned}
\end{equation*}
\endgroup
and $\mathcal{L}_{rel}$ is defined analogously on relative camera translations and rotations for valid image pairs. Here, $r_i$, $t_i$ and $\hat{r}_i$, $\hat{t}_i$ denote the ground truth and predicted camera translation and normalized rotation vector for camera $i$, and $\beta_r=0.1$.

We optimize Eq.~(\ref{equation:stage_1}) with AdamW for 80k iterations, a cosine learning-rate schedule with peak \(10^{-4}\), and 4k warmup iterations. Each batch samples one scene and 2, 3, 5, or 8 images per example such that the views form a single pose graph with co-visibility greater than \(20\%\). A linear view-count scheduler shifts from 2 and 3-view batches early in training to 5 and 8-view batches later. We use random horizontal rotation augmentation and clip gradients to norm 1.0.

\paragraph{Layout semantic training.}
We train the floor-layout semantic branch from scratch while freezing the geometry branch and reusing its features. We use the same data-sampling schedule and a similar learning-rate schedule, with maximum learning rate \(3.5 \times 10^{-4}\). The semantic branch uses dense \textit{MOIS} supervision with a factorized objective: wall-presence prediction for each anchor-column pair, and ordered wall-slot classification on positives.

Let \(m \in \{1,\dots,M\}\) index sampled anchor columns, with anchor \(m\) corresponding to \((i_m,c_m)\). For anchor \(m\) and column \((j,d)\), the decoder produces dense \textit{MOIS} probabilities \(\hat{\mathcal{S}}_{m,j,d,r}=\sigma(z^{\mathrm{cls}}_{m,j,d,r})\), where \(r \in \{0,\dots,K-1\}\) indexes the ordered wall slot relative to the anchor. The binary supervision is \(\mathcal{S}^{\mathrm{gt}}_{m,j,d,r}\in\{0,1\}\).

We aggregate wall-slot logits into a presence logit \(z_{m,j,d}^{\mathrm{pres}}=\log \sum_{r=0}^{K-1}\exp(z_{m,j,d,r}^{\mathrm{cls}})\), with target \(y_{m,j,d}^{\mathrm{pres}} \in \{0,1\}\) indicating whether \((j,d)\) is assigned to any ordered wall slot of anchor \(m\). The wall-presence and ordered-slot classification losses are
\begingroup
\small
\begin{equation*}
\begin{aligned}
\mathcal{L}_{\mathrm{pres}} &=
\frac{1}{|\Omega_{\mathrm{valid}}|}
\sum_{(m,j,d)\in\Omega_{\mathrm{valid}}}
\mathrm{BinaryCrossEntropy}\!\left(z_{m,j,d}^{\mathrm{pres}},\, y_{m,j,d}^{\mathrm{pres}}\right), \\
\mathcal{L}_{\mathrm{cls}} &=
\frac{1}{|\Omega_{\mathrm{pos}}|}
\sum_{(m,j,d)\in\Omega_{\mathrm{pos}}}
\mathrm{CrossEntropy}\!\left(z_{m,j,d,:}^{\mathrm{cls}},\, y_{m,j,d}^{\mathrm{cls}}\right),\\
\mathcal{L}_{\textit{MOIS}} &= \mathcal{L}_{\mathrm{pres}}+\lambda_{\mathrm{cls}}\mathcal{L}_{\mathrm{cls}}.
\end{aligned}
\end{equation*}
\endgroup
Here, \(\Omega_{\mathrm{valid}}\) is the set of valid projected panorama columns after masking invalid regions; \(\Omega_{\mathrm{pos}}\subseteq\Omega_{\mathrm{valid}}\) contains columns assigned to one of the \(K\) ordered slots, with label \(y_{m,j,d}^{\mathrm{cls}}\in\{0,\dots,K-1\}\).

\endgroup

\section{\textit{Layout Chaining} Post Processing}
\label{sec:lc_alg}

\textit{Layout Chaining} (\textit{LC}) decodes the dense \textit{MOIS} tensor \(\hat{\mathcal{S}}\) into compact structured layout outputs: global column-to-wall assignments \(\mathcal{M}\), room-level wall sequences \(\{\mathcal{L}^m\}\), and metric wall endpoints \(\{V^m\}\). The input is dense because every anchor column predicts cross-view masks for the current wall and neighboring wall slots, creating redundant local evidence for wall identity and adjacency.

\textit{LC} reduces this evidence in four stages: 
(1) same-view current-wall masks are grouped into per-view wall instances \(L_n\); 
(2) cross-view same-wall evidence merges them into global wall instances, defining \(\mathcal{M}\); 
(3) neighboring-wall slots induce directed within-room adjacency and room-level wall sequences \(\{\mathcal{L}^m\}\); and 
(4) predicted point maps and wall-line parameters provide metric support for wall-line fitting and endpoint recovery, yielding \(\{V^m\}\).

\begin{algorithm}[H]
\small
\SetAlgoNlRelativeSize{0}
\SetAlCapSkip{1pt}
\SetAlgoInsideSkip{1pt}
\caption{\textit{MvFFN} Post-processing: \textit{Layout Chaining}}
\label{alg:layout_chaining}
\LinesNumbered
\KwIn{Dense \textit{MOIS} field \(\hat{\mathcal{S}} \in [0,1]^{N \times W \times N \times W \times K}\), floor point maps \(\{\hat{\mathcal{P}}_{i,c}\}\), and wall-line parameters \(\{\hat{\lambda}_{i,c}\}\)}
\KwOut{Column-to-wall assignments \(\mathcal{M}\), within-room connectivity \(\{\mathcal{L}^m\}\), and room endpoints \(\{V^m\}\)}

\tcc{Stage 1: Per-view wall instances}
\For{each view \(n\)}{
  Define per-view current wall segmentation: \(\hat{\mathcal{S}}^{\mathrm{view}}_n = \hat{\mathcal{S}}[n,:,n,:,0]\)\;
  Define per-view column nodes: \(\mathcal{V}^{\mathrm{view}}_n=\{1,\dots,W\}\)\;
  \For{column pair \((c_1,c_2)\in \mathcal{V}^{\mathrm{view}}_n \times \mathcal{V}^{\mathrm{view}}_n\)}{
    Define same-wall column edge: add \((c_1,c_2)\) to \(\mathcal{E}^{\mathrm{view}}_n\) if \(\mathrm{IoU}(\hat{\mathcal{S}}^{\mathrm{view}}_n[c_1],\hat{\mathcal{S}}^{\mathrm{view}}_n[c_2])>\tau_{\mathrm{iou}}\)\;
  }
  Build per-view same-wall column graph: \(\mathcal{G}^{\mathrm{view}}_n=(\mathcal{V}^{\mathrm{view}}_n,\mathcal{E}^{\mathrm{view}}_n)\)\;
  Collect per-view wall instances: \(L_n = \textbf{ConnectedComponents}(\mathcal{G}^{\mathrm{view}}_n)\)\;
  \For{each connected component \(l \in L_n\)}{
    Select anchor column: \(c^{*} = \arg\max_{c \in l} \deg(c)\)\;
    Define instance mask: \(l^{*} = \hat{\mathcal{S}}^{\mathrm{view}}_n[c^{*}]\)\;
  }
}

\tcc{Stage 2: Cross-view global wall instances}
Define cross-view wall nodes: \(\mathcal{V}_{\mathrm{cross}}=\bigcup_{n=1}^N L_n\)\;
\For{instance pair \((l_i,l_j)\in L_{n_1} \times L_{n_2},\; n_1 \neq n_2\)}{
  Define cross-view same-wall instance edge: add \((l_i,l_j)\) to \(\mathcal{E}_{\mathrm{cross}}\) if \(\mathrm{IoU}(l_i^*,l_j^*)>\tau_{\mathrm{iou}}\)\;
}
Build cross-view same-wall instance graph: \(\mathcal{G}_{\mathrm{cross}}=(\mathcal{V}_{\mathrm{cross}},\mathcal{E}_{\mathrm{cross}})\)\;
Collect global wall instances via cross-view column-to-wall assignments: \(\mathcal{M}=\textbf{ConnectedComponents}(\mathcal{G}_{\mathrm{cross}})\)\;

\tcc{Stage 3: Room connectivity and partial layouts}
Define room-wall nodes: \(\{w^\ell\}\), induced by \(\mathcal{M}\)\;
Define directed room-wall edges: \(\mathcal{E}_{\mathrm{room}}=\{(w_i,w_j)\mid \text{the } k\!\rightarrow\!0 \text{ predictions in } \hat{\mathcal{S}}[:,:,:,:,1:K\!-\!1] \text{ support } w_i \rightarrow w_j\}\)\;
Build directed room graph: \(\mathcal{G}_{\mathrm{room}}=(\{w^\ell\},\mathcal{E}_{\mathrm{room}})\)\;
Filter invalid edges by geometry: \(\mathcal{G}_{\mathrm{room}}=\textbf{GeomFilter}(\mathcal{G}_{\mathrm{room}})\)\;
\For{each weakly connected component \(\mathcal{H}^m\subseteq \mathcal{G}_{\mathrm{room}}\)}{
  Collect within-room connectivity: \(\mathcal{L}^m = \textbf{LongestPath}(\mathcal{H}^m)\)\;
}

\tcc{Stage 4: Final wall geometry}
Define global wall parameters: \(\{(\hat{b}^\ell,\hat{\mathbf{v}}^\ell)\}=\textbf{LineFit}(\{\hat{\mathcal{P}}_{i,c}\},\mathcal{M})\)\;
\For{each partial room layout \(\mathcal{L}^m\)}{
  Define room endpoints: \(V^m=(v^{m,1},\dots,v^{m,K_m})=\textbf{LineIntersect}(\mathcal{L}^m,\mathcal{M})\)\;
}
\end{algorithm}

\section{Intermediate \textit{MOIS} Prediction Evaluation}
\label{sec:mois_eval}

\paragraph{\textit{MOIS} evaluation.}
We evaluate the intermediate \textit{MOIS} representation before \textit{Layout Chaining} with \textit{Pooled Ordered Wall-Slot IoU}, an $IoU$ metric computed separately for each ordered \textit{MOIS} wall slot. For each \(N\)-view test tuple, we sample anchor columns from the conditioning panoramas and retain anchors with valid ordered slot targets. Each anchor instantiates the cross-view slot masks defined in Sec.~\ref{sec:mois} of the main paper. For each slot, we accumulate intersections and unions between the predicted and ground-truth masks over all valid anchors, views, and panorama columns before taking the ratio. This pixel-weighted pooling measures aggregate cross-view mask quality and avoids the high variance of query-averaged \(\mathrm{mIoU}\) for small or weakly visible wall instances.

\paragraph{\textit{LGT-Net} + \textit{CovisPose} baseline.}
No prior method directly predicts ordered cross-view wall instances, so we construct a composed baseline from \textit{LGT-Net}~\cite{jiang2022lgtnet} and \textit{CovisPose}~\cite{hutchcroft2022covispose}. The goal is to approximate \textit{MOIS} by first assigning each anchor column to an ordered set of local wall slots in its source view, then propagating those slots across views. Given the single-view layout predicted by \textit{LGT-Net}, we convert it into a floor-plan polygon whose edges define the local slots: the edge intersected by the anchor ray is slot 1, and the next two adjacent edges in clockwise order are slots 2 and 3. \textit{CovisPose} then provides cross-view column correspondences, which propagate these source-view slots to other panoramas and produce cross-view slot masks. We evaluate these masks against the same ordered wall-instance targets using \textit{Pooled Ordered Wall-Slot IoU}.

\begin{table}[h]
\centering
\caption{
\textit{Pooled Ordered Wall-Slot IoU}$\uparrow$ (\%) for raw \textit{MOIS} predictions before \textit{Layout Chaining}. Wall 1 denotes the anchor wall, and 2-3 denote the next ordered neighboring walls.
}
\label{tab:mois_zind}
\resizebox{0.65\linewidth}{!}{
\begin{tabular}{c|ccc|ccc}
\hline
                        & \multicolumn{3}{c|}{5-view inputs}            & \multicolumn{3}{c}{8-view inputs}             \\ \cline{2-7} 
Method                  & wall 1        & wall 2        & wall 3        & wall 1        & wall 2        & wall 3        \\ \hline
\textit{\textbf{MvFFN}} & \textbf{81.9} & \textbf{71.1} & \textbf{57.8} & \textbf{75.4} & \textbf{63.8} & \textbf{47.7} \\
\textit{LGT} + \textit{CP}                & 63.7          & 37.8          & 20.3          & 60.4          & 34.5          & 18.5          \\ \hline
\end{tabular}
}

\end{table}

Table~\ref{tab:mois_zind} shows that \textit{MvFFN} substantially improves raw \textit{MOIS} prediction quality before any \textit{Layout Chaining} post-processing. Wall 1 mainly reflects cross-view wall-instance recovery, while walls 2 and 3 evaluate the ordered neighboring-wall predictions that encode local within-room connectivity. The larger gains on walls 2 and 3 show that \textit{MvFFN} learns not only same-wall correspondence but also adjacent-wall structure more reliably than the composed \textit{LGT-Net}+\textit{CovisPose} baseline. These stronger local predictions provide better inputs to \textit{Layout Chaining}, supporting the downstream gains in final wall and junction accuracy.

\section{Experiments on HM3D-XV Dataset}
\label{sec:hm3d_exp}

We use \texttt{HM3D-XV} as an auxiliary cross-domain validation benchmark to evaluate whether \textit{MvFFN} and the \textit{MOIS} formulation remain effective beyond \texttt{ZInD-XV}. \texttt{HM3D-XV} is derived from residential \texttt{HM3DSem} scenes~\cite{yadav2023hm3dsem} with sufficient mesh coverage, and contains 17,943 panoramas and 372 floors from 180 scenes. Compared with \texttt{ZInD-XV}, which contains 2,706 floors, \texttt{HM3D-XV} is smaller and noisier, contains furnished mesh-rendered scenes with rendering artifacts, and has sparser column-level supervision, with 27.3\% of image columns masked after annotation cleanup. Dataset construction details are provided in Sec.~\ref{sec:hm3d-pipeline}. This makes \texttt{HM3D-XV} a challenging setting for both pose estimation and layout reconstruction. We therefore use \texttt{ZInD-XV} as an anchor dataset and finetune \textit{MvFFN} with mixed \texttt{ZInD-XV}/\texttt{HM3D-XV} samples to test whether the learned geometry and \textit{MOIS} representations remain useful under domain shift and noisy annotations.

For evaluation, we construct \(N\)-view test splits from per-floor tuples generated by covisibility-based sampling. Each floor receives a sample count proportional to its number of panoramas, capped by the number of valid tuples; remaining samples are redistributed according to residual capacity. This yields 951 tuples for the 5-view split and 991 tuples for the 8-view split, spanning all floors with valid \(N\)-view connected components while keeping floor-level representation roughly proportional to scene size. Both pose and layout experiments use these splits. Because some comparisons rely on \textit{CovisPose}-derived priors, we additionally report those results on the \textit{CovisPose}-feasible subset defined in Sec.~\ref{sec:hm3d_pose}. Sec.~\ref{sec:hm3d_pose} evaluates pose estimation, and Sec.~\ref{sec:hm3d_layout} evaluates floor layout reconstruction.

\subsection{Camera Pose Estimation}

\label{sec:hm3d_pose}

We follow the same pose-evaluation protocol as the \texttt{ZInD-XV} experiments and report pose accuracy on a \textit{CovisPose}-feasible subset of \texttt{HM3D-XV}. Starting from the \texttt{ZInD-XV}-trained model, we fine-tune the geometry branch on \texttt{HM3D-XV} using the same losses, view-count schedule, and dataloader setup as Sec.~\ref{sec:geometry_training}, with a 60k-iteration learning-rate schedule and the 40k-iteration checkpoint used for evaluation. Training batches mix \texttt{HM3D-XV} and \texttt{ZInD-XV} examples at a 1:1 ratio to adapt to \texttt{HM3D-XV} while retaining broader geometric variation from \texttt{ZInD-XV}.

Furnished, mesh-rendered \texttt{HM3D-XV} panoramas differ substantially from the \texttt{ZInD} panoramas used to train the available \textit{CovisPose} checkpoint. As a result, some pairwise \textit{CovisPose} estimates are unreliable. We filter pairs with rotation error above \(30^\circ\) or translation error above 2\,m, and retain only tuples whose remaining estimates form a connected \textit{GST} pose graph. This removes invalid pairwise links that would otherwise dominate pose and layout metrics, retaining approximately 60\% of the 5-view tuples and 65\% of the 8-view tuples. We compare \textit{CovisPose}+\textit{GST}, \textit{VGGT}, and \textit{MvFFN}. For \textit{MvFFN}, we use the same rotation-prior construction as in \texttt{ZInD-XV}, with \textit{CovisPose} rotations refined by vanishing-angle correction. For the correspondence attention mask, we use ground-truth column-patch correspondences to provide a reliable inductive bias for sparse multi-view training on \texttt{HM3D-XV}.

\begin{table}[h]
\centering
\caption{Camera pose evaluation on the \textit{CovisPose}-feasible subset of \texttt{HM3D-XV}. Mn/Std/Med denote mean, standard deviation, and median errors.
}
\label{tab:pose_hm3d}
\setlength{\tabcolsep}{3pt}
\renewcommand{\arraystretch}{1.12}
\resizebox{0.8\linewidth}{!}{
\begin{tabular}{c|cccccc|cccccc}
\hline
\multirow{2}{*}{Methods}        & \multicolumn{3}{c}{5v Pos (cm)$\downarrow$}   & \multicolumn{3}{c|}{5v Rot ($^\circ$)$\downarrow$} & \multicolumn{3}{c}{8v Pos (cm)$\downarrow$}   & \multicolumn{3}{c}{8v Rot ($^\circ$)$\downarrow$} \\ \cline{2-13} 
                                & Mn            & Std           & Med           & Mn              & Std             & Med            & Mn            & Std           & Med           & Mn              & Std            & Med            \\ \hline
\textit{VGGT}                   & 35.2          & 109.3         & \textbf{10.4} & 6.1             & 16.3            & 2.1            & 62.3          & 71.8          & 37.4          & 13.7            & 6.4            & 12.3           \\
\textit{CovisPose}+\textit{GST} & 30.1          & 24.7          & 19.3          & 2.2             & 3.3             & \textbf{1.1}   & 44.0          & 59.7          & 25.9          & 2.9             & 4.4            & \textbf{1.4}   \\
\textit{MvFFN}                  & \textbf{26.8} & \textbf{21.2} & 21.5          & \textbf{1.8}    & \textbf{0.6}    & 1.6            & \textbf{32.0} & \textbf{23.5} & \textbf{26.6} & \textbf{1.9}    & \textbf{0.5}   & 1.7            \\ \hline
\end{tabular}
}
\end{table}

As shown in Table~\ref{tab:pose_hm3d}, under reliable correspondence initialization, \textit{MvFFN} produces stable pose estimates on the \textit{CovisPose}-feasible subset of \texttt{HM3D-XV}. Compared with \textit{CovisPose}+\textit{GST}, it achieves lower mean errors and lower variance across both 5-view and 8-view settings, while \textit{CovisPose}+\textit{GST} and \textit{VGGT} obtain lower medians in some cases. These results suggest that the \textit{MvFFN} geometry branch, under this setting, can support stable multi-view pose estimation in the furnished \texttt{HM3D-XV} domain.

\subsection{Layout Estimation}
\label{sec:hm3d_layout}


For layout estimation, our goal is to test whether the learned \textit{MOIS} representation and layout decoder can support multi-room reconstruction in the furnished, mesh-rendered \texttt{HM3D-XV} domain. Following the pose evaluation, we use the same prior setup for \textit{MvFFN} as in Sec.~\ref{sec:hm3d_pose}. Starting from the \texttt{HM3D-XV} tuned geometry checkpoint, we retrain only the floor-layout semantic branch with mixed \texttt{ZInD-XV} and \texttt{HM3D-XV} samples, using a 2:1 \texttt{ZInD-XV}: \texttt{HM3D-XV} sampling ratio and otherwise keeping the training settings the same as Sec.~\ref{sec:training_details}. We compare against \textit{LGT-Net} based modular baselines using either \textit{CovisPose} or ground truth camera poses. As in Sec.~\ref{sec:layout_evals} of the main paper, the modular pipeline separates panorama-level boundary prediction from layout vectorization: \textit{LGT-Net} (\textit{LGT}) provides floor boundary predictions, and \textit{LGT-Net} post-processing (\textit{LGT-PP}) converts them into vectorized layouts. \textit{RoomFormer} and \textit{PolyRoom} are omitted because they are substantially weaker on the main \texttt{ZInD-XV} layout benchmark. 


\begin{table}[!ht]
\centering
\vspace{0pt}
\caption{Floor-layout evaluation on \texttt{HM3D-XV}. Distance is in cm. Geom. denotes the intermediate geometric input and Layout denotes the vectorization backend. GT-pose denotes a control group.}
\label{tab:floor_layout_hm3d}
\vspace{0pt}
\setlength{\tabcolsep}{2pt}
\renewcommand{\arraystretch}{1.12}

\resizebox{1.0\linewidth}{!}{%
\begin{tabular}{ccc|cccc|cccc|cccc|cccc}
\hline
\multicolumn{1}{c|}{\multirow{2}{*}{pose}} & \multicolumn{1}{c|}{\multirow{2}{*}{pts}} & \multirow{2}{*}{layouts} & \multicolumn{4}{c|}{5-view wall}                              & \multicolumn{4}{c|}{5-view junction}                          & \multicolumn{4}{c|}{8-view wall}                              & \multicolumn{4}{c}{8-view junction}                           \\
\multicolumn{1}{c|}{}                      & \multicolumn{1}{c|}{}                     &                          & prec          & recall        & f1            & dist          & prec          & recall        & f1            & dist          & prec          & recall        & f1            & dist          & prec          & recall        & f1            & dist          \\ \hline
\multicolumn{1}{c|}{\textit{CP}}           & \multicolumn{1}{c|}{\textit{LGT}}         & \textit{LGT}             & 57.0          & 62.6          & 59.7          & 51.8          & 18.2          & 33.3          & 23.5          & 53.1          & 59.4          & 60.2          & 59.8          & 52.1          & 19.0          & 30.5          & 23.4          & 53.8          \\ \cline{1-3}
\multicolumn{1}{c|}{GT}                    & \multicolumn{1}{c|}{\textit{LGT}}         & \textit{LGT}             & 69.1          & 66.8          & 67.9          & 36.9          & 25.5          & 44.3          & 32.4          & 35.7          & 69.7          & 64.6          & 67.1          & 39.9          & 25.5          & 38.9          & 30.8          & 38.6          \\ \cline{1-3}
\multicolumn{3}{c|}{\textit{\textbf{MvFFN}}}                                                                      & \textbf{77.8} & \textbf{69.6} & \textbf{73.5} & \textbf{22.1} & \textbf{55.9} & \textbf{38.8} & \textbf{45.8} & \textbf{24.1} & \textbf{75.0} & \textbf{71.6} & \textbf{73.3} & \textbf{28.9} & \textbf{52.9} & \textbf{35.0} & \textbf{42.1} & \textbf{29.3} \\ \hline
\end{tabular}
}

\vspace{-4pt}
\end{table}

As shown in Table~\ref{tab:floor_layout_hm3d}, under the reliable prior setup, \textit{MvFFN} substantially improves layout accuracy on \texttt{HM3D-XV} despite the smaller and noisier training set. It achieves much higher wall precision and \(F1\) than the \textit{LGT-Net} based baselines, and reduces matched wall distance in both 5-view and 8-view settings. The gains are especially large at the junction level, where the modular baselines remain weak even with ground truth poses. This suggests that \textit{MOIS} remains an effective supervision target under sparse column-level annotations, and helps \textit{MvFFN} recover wall connectivity and junction structure more reliably than geometry-first modular pipelines.

\section{More Ablation Studies}
\label{sec:more_ablation}

The main paper reports the architecture ablation for the two-branch, two-stage \textit{MvFFN} design. Here we provide two additional ablations: Sec.~\ref{sec:inductive_bias_ablation} studies the rotation and correspondence priors used for sparse multi-view training, and Sec.~\ref{sec:mois_window_ablation} evaluates larger \textit{MOIS} prediction windows to test the effect of local prediction length.

\subsection{Inductive Bias}
\label{sec:inductive_bias_ablation}

\begin{figure}[H]
\centering
    \includegraphics[width=1.0\linewidth]{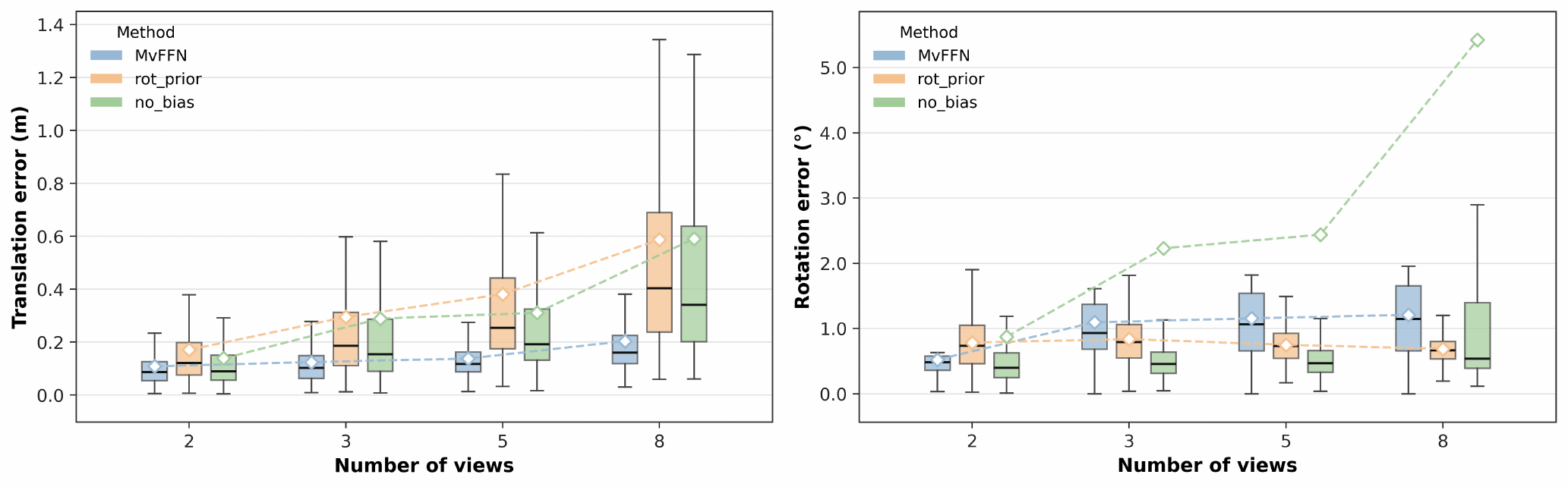}
    \vspace{-16pt}
    \caption{Effect of inductive bias across input views. Box plots show translation (left) and rotation (right) errors under different inductive-bias variants. Dashed lines and diamond markers indicate mean errors. Stronger inductive bias improves translation accuracy and suppresses extreme rotation outliers.} 
    \label{fig:ablation_inductive_bias}
    \vspace{-14pt}
\end{figure}

We evaluate \textit{MvFFN} across varying numbers of input panoramas, using three levels of inductive bias described in Sec.~\ref{sec:inductive_bias} of the main paper: no inductive bias, rotation priors only, and rotation priors with correspondence-based attention masks. As shown in Fig.~\ref{fig:ablation_inductive_bias}, stronger inductive bias has its clearest effect on translation accuracy: the full \textit{MvFFN} maintains substantially lower translation error across all input-view settings, while the weaker-bias variants degrade much more rapidly as the number of views increases. For rotation, the main benefit is robustness rather than uniformly lower central statistics. In some settings, the no-bias variant can appear slightly better in the median, but this occurs only near the center of the distribution, where all methods already achieve low angular error. Its mean is nonetheless much higher because the error distribution develops a heavy tail: in the 8-view setting, the no-bias variant reaches $17.5^\circ$ at p90 and $27.8^\circ$ at p95, whereas the full \textit{MvFFN} remains at $2.5^\circ$ at p95. Thus, the proposed inductive bias mainly suppresses large rotation failures and produces much more stable estimates in hard cases. Since all variants already use vanishing-angle priors, this gain is not explained by rotation snapping alone, but by the additional regularization introduced by the multi-view inductive bias.

\subsection{\textit{MOIS} Prediction Window}
\label{sec:mois_window_ablation}
We train two additional \textit{MvFFN} variants with larger \textit{MOIS} prediction windows of 5 and 7 ordered walls, and report the results in Table~\ref{tab:ablation_walls}. For a controlled comparison, \textit{Layout Chaining} is kept unchanged and uses only the first three predicted wall slots for all variants, so the ablation isolates the effect of training \textit{MvFFN} to predict longer \textit{MOIS} windows. Expanding the prediction window leads to a consistent drop in layout accuracy, with 5-view wall \(F1\) decreasing from 88.1 to 85.6 and 84.4, and 8-view junction \(F1\) decreasing from 75.0 to 71.3 and 67.1 for the 5-wall and 7-wall variants, respectively. Matched-wall distance also worsens in the 8-view setting. These results support our choice of a 3-wall \textit{MOIS} window: a compact local prediction target is easier to learn under sparse wide-baseline observations, while longer ordered wall sequences accumulate more ambiguity from occlusion, room-boundary uncertainty, and distant structural details. The degradation is gradual rather than catastrophic: increasing the \textit{MOIS} window makes the task harder, but does not break the formulation. This suggests that locality is helpful, with 3 walls giving the best operating point in our setting, while the broader advantage comes from replacing direct global prediction with an anchor-conditioned local representation.

\begin{table}[!ht]
\centering
\vspace{-8pt}
\caption{\textit{MvFFN} layout accuracy reported on 5- and 8-view test set from \texttt{ZInD-XV}. The model is trained to predict different numbers of adjacent walls in \textit{MOIS} prediction.}
\label{tab:ablation_walls}
\vspace{-6pt}
\setlength{\tabcolsep}{2pt}
\renewcommand{\arraystretch}{1.12}

\resizebox{1.0\linewidth}{!}{%
\begin{tabular}{c|cccc|cccc|cccc|cccc}
\hline
\multirow{2}{*}{\# walls} & \multicolumn{4}{c|}{5-view wall}                              & \multicolumn{4}{c|}{5-view junction}                          & \multicolumn{4}{c|}{8-view wall}                              & \multicolumn{4}{c}{8-view junction}                           \\
                          & Prec$\uparrow$          & Rec$\uparrow$        & $F1\uparrow$            & Dist$\downarrow$          & Prec$\uparrow$          & Rec$\uparrow$        & $F1\uparrow$            & Dist$\downarrow$          & Prec$\uparrow$          & Rec$\uparrow$        & $F1\uparrow$            & Dist$\downarrow$          & Prec$\uparrow$          & Rec$\uparrow$        & $F1\uparrow$            & Dist$\downarrow$          \\ \hline
3-wall                    & 88.5          & \textbf{87.7} & \textbf{88.1} & \textbf{13.9} & \textbf{79.2} & \textbf{68.2} & \textbf{73.3} & \textbf{13.8} & \textbf{89.4} & \textbf{87.5} & \textbf{88.4} & \textbf{19.0} & \textbf{81.0} & \textbf{69.8} & \textbf{75.0} & \textbf{19.1} \\
5-wall                    & \textbf{88.7} & 82.8          & 85.6          & 14.7          & 78.8          & 64.2          & 70.8          & 17.0          & 88.4          & 83.7          & 86.0          & 19.6          & 79.4          & 64.7          & 71.3          & 23.2          \\
7-wall                    & 87.6          & 81.4          & 84.4          & 16.2          & 76.3          & 58.7          & 66.4          & 18.7          & 87.7          & 78.9          & 83.1          & 19.2          & 77.9          & 59.0          & 67.1          & 21.7          \\ \hline
\end{tabular}
}

\vspace{-16pt}
\end{table}

\subsection{Architecture Ablation Details}
We compare four architecture variants. The shared-encoder model removes the dedicated semantic feature encoder and predicts \textit{MOIS} directly from geometry-branch features. The no-fusion model keeps separate geometry and semantic branches but removes geometry-to-semantic feature fusion, while retaining the same downstream \textit{Layout Chaining} procedure. The one-stage model uses the full two-branch architecture but trains all objectives jointly from scratch. The full model uses separate geometry and semantic branches with the proposed two-stage schedule.

The no-fusion variant substantially degrades layout recall, \(F1\), and matched-wall distance, indicating that semantic image features alone are insufficient and that geometry-conditioned semantic features are important for cross-view wall grouping and within-room connectivity. The shared-encoder and one-stage variants both become unstable during training: the shared-encoder model collapses after roughly 25k steps, while the one-stage model collapses after roughly 35k steps. This suggests conflicting optimization pressure between dense geometry prediction, which relies on precise one-to-one cross-view matching, and \textit{MOIS} prediction, which groups columns by wall identity and room structure. Separating the geometry and semantic branches and training them in stages improves both stability and final layout accuracy.

\section{Dataset Preparation}
\label{sec:dataset_prep}

\subsection{ZInD-XV Dataset Preparation}
\label{sec:zind-pipeline}

\texttt{ZInD-XV} is derived from all scenes in the \texttt{Zillow Indoor Dataset} (\texttt{ZInD})~\cite{cruz2021zillow}. It contains 66,834 panoramas, 167,983 unique wall instances, 23,651 rooms, and 2,706 floors from 1,524 residential house scenes in North America. \texttt{ZInD-XV} retains the original panoramas and room layouts in the global frame, but augments them with globally indexed multi-room floor-plan annotations. Specifically, each floor is represented as a set of room layouts with doors and windows, encoded as closed room polygons over globally unique wall instances in the same format as \texttt{FloorPlan60K} used in~\cite{li2025badgr}. Because \texttt{ZInD} often partitions large open spaces into smaller partial rooms, we recursively merge adjacent rooms connected by openings that do not resemble doorways. As a result, the room partition in \texttt{ZInD-XV} is less fragmented than the original \texttt{ZInD} partition in open-plan areas, while still preserving meaningful room-level structure rather than collapsing each floor into a single geometric region. These global wall instances are further projected into each panorama to produce column-wise floor boundaries and wall-instance semantic labels.

Because a single panorama can observe walls from multiple rooms, and visible room connectivity depends on doorway state and view position, these per-panorama labels cannot be obtained by projecting a single room polygon independently. Instead, for each panorama, we search the closest room graph connected through open doors and select the room configuration whose rendered floor boundary best matches the 1D floor boundary predicted by \textit{CovisPose}~\cite{hutchcroft2022covispose}. This graph is then used to generate the per-panorama floor-boundary and wall-instance labels. Finally, we apply a manual cleanup pass to mask panorama columns with inconsistent floor boundaries or incorrect wall-instance assignments. We plan to release the \texttt{ZInD-XV} annotations, including global floor-plans with unique wall instances, per-panorama floor boundaries, wall-instance semantic labels, and validity masks.

\subsection{HM3D-XV Dataset Preparation}
\label{sec:hm3d-pipeline}
\label{sec:hm3d_dataset}

\texttt{HM3D-XV} is constructed from residential scenes in the \texttt{Habitat--Matterport 3D Semantic Dataset} (\texttt{HM3DSem})~\cite{yadav2023hm3dsem}, restricted to scenes with sufficient mesh coverage to better match the residential scene distribution of \texttt{ZInD-XV}. It contains 17,943 panoramas from 372 eligible floors across 180 scenes, split into 287 training floors and 85 validation floors following \texttt{HM3DSem}. Although \texttt{HM3DSem} provides textured house meshes with semantic labels, it does not provide the panorama-based annotations required by our model: camera-posed panoramas, globally indexed wall instances, room-level wall sequences, and per-panorama column-wise floor-boundary and wall-instance labels. We construct these annotations in three stages: automated floor preparation, room-level wall sequence annotation, and per-panorama column-wise annotation. We plan to release the \texttt{HM3D-XV} annotations, including global floor-plans with unique wall instances, per-panorama camera poses, floor boundaries, wall-instance semantic labels, and validity masks.


\paragraph{Automated floor preparation.}
This stage converts each multi-floor \texttt{HM3DSem} scene mesh into per-floor sub-scenes with panoramas of known camera poses and initialized global wall instances. We first segment each scene mesh by floor using the semantic labels to identify floor and ceiling geometry. The floor and ceiling heights used for this partition are estimated from the median vertical positions of these surfaces, and nearby split floor regions are merged with simple geometric heuristics. For each floor, we sample panorama centers from the Habitat navigation mesh with a density-controlled greedy Poisson-disk strategy~\cite{bridson2007_poisson_sampling}, targeting approximately 2 panoramas per square meter at a fixed height of 1.2\,m. Using the \textit{Habitat-Sim} renderer~\cite{ramakrishnan2021habitat}, we render $2048\times1024$ RGB equirectangular panoramas from the textured mesh with upright camera orientation. Candidate samples that fall on clutter are discarded, and panoramas with severe visible mesh defects are removed during manual inspection. We then initialize global wall instances at the floor level. Vertical wall meshes are grouped by semantic wall instance label from \texttt{HM3DSem} and projected to the floor plane, and a 2D line is fit to each projected wall instance with \textit{RANSAC} to recover the wall parameters defined in~\cite{li2025badgr}.

\paragraph{Room-level wall sequence annotation.}
Starting from the initialized global wall instances, this step produces the room-level wall sequences used as the final multi-room floor-plan layout for each floor. We designed a custom graphical interface that lets annotators reorder, merge, and split wall instances, and group them into room-wise ordered sequences with globally consistent wall identities. Each room sequence may be closed or open, since incomplete mesh coverage in \texttt{HM3DSem} often prevents some rooms from being represented as closed polygons.

\paragraph{Per-panorama column-wise annotation.}
Given the annotated multi-room wall layout and panorama camera poses, we project floor geometry and wall instances into each panorama to generate column-wise floor boundaries and wall-instance semantic labels. A separate manual annotation pass is then used to mask out columns with subpar projected assets, for example when floor boundaries are misaligned with visible structure or wall-instance assignments are incorrect.

\section{Baseline Details}
\label{sec:baseline_details}

We evaluate modular baselines under a common three-stage pipeline: (1) pose estimation in a shared floor-plane frame, (2) construction of pose-aligned floor boundaries or point-maps, and (3) layout vectorization. This setup lets us compare existing pose, boundary, and layout components under the same sparse multi-view input setting. Unlike \textit{MvFFN}, these baselines do not perform joint multi-view reasoning directly from panorama images in the final layout stage; the layout backend receives only pose-aligned geometric inputs, so upstream pose or boundary errors are already fixed before layout prediction.

\subsection{Pose Evaluation}
\paragraph{Pose estimation baselines.}
We compare against two pose baselines: \textit{CovisPose}+\textit{Greedy Spanning Tree} (\textit{GST}) and \textit{VGGT}. For \textit{CovisPose}, pairwise relative panorama poses are predicted for all panorama pairs, and global 3-DoF poses (2D position and 1D rotation) are recovered by chaining relative transforms along a co-visibility-ranked greedy spanning tree, following prior panoramic pose-estimation work~\cite{nejatishahidin2023graph,li2025badgr}. \textit{VGGT}~\cite{wang2025vggt} does not operate directly on \(360^\circ\) panoramas, so we adapt it through perspective decomposition. We exclude \textit{GraphCovis}~\cite{nejatishahidin2023graph} and \textit{PanoPose}~\cite{tu2024panopose} because public code is unavailable.

For each view-count setting, we evaluate on 1,000 \(N\)-view test tuples from \texttt{ZInD-XV}. Tuples are generated following the prior pose-evaluation protocol~\cite{nejatishahidin2023graph}: views are sampled from the same floor with minimum pairwise co-visibility of 0.3, and floors are sampled to maintain broad floor coverage. Predicted pose graphs are aligned to ground truth before evaluation following~\cite{nejatishahidin2023graph,li2025badgr}. We report per-panorama position error in centimeters and yaw error in degrees, summarized by mean, standard deviation, and median over all evaluated panoramas.

\paragraph{VGGT panorama adaptation.}
Since \textit{VGGT} operates on perspective images, each \(360^\circ\) panorama is decomposed into overlapping rectilinear crops with \(120^\circ\) field of view, \(60^\circ\) horizontal overlap, resolution \(560\times560\), and pitch angles \(0^\circ\) and \(-30^\circ\). All crops from all panoramas are processed jointly in a single forward pass. We recover metric scale by aligning the vertical histogram peak of the predicted point cloud to the camera height provided by \texttt{ZInD}~\cite{cruz2021zillow}. We then estimate one planar pose per panorama by compensating each crop pose for its known azimuthal offset and robustly fitting a consensus position and yaw across crops.

\subsection{Layout Baseline Pipelines}
\paragraph{Layout baseline pipelines.}
Given Stage-1 poses, all modular layout baselines reconstruct multi-view layouts by first building geometric inputs in the global floor frame and then applying a layout vectorization backend. Unless otherwise noted, we use \textit{CovisPose} poses; GT-pose variants replace them with ground-truth poses to isolate layout error from pose error. \textit{LGT-Net} provides panorama-level floor-wall boundary cues, while \textit{RoomFormer} and \textit{PolyRoom} use fused top-down geometric inputs. We do not use \textit{VGGT} poses for layout baselines because its pose accuracy is substantially worse than \textit{CovisPose} on \texttt{ZInD-XV}.

\paragraph{LGT-Net backend.}
\textit{LGT-Net}~\cite{jiang2022lgtnet} is applied independently to each panorama using the pretrained \texttt{ZInD} checkpoint without fine-tuning. Its predicted floor-wall boundary is converted into a local floor-plan polygon using Manhattan-world post-processing with \textit{Douglas--Peucker} simplification \((\epsilon=0.1)\), then projected into the global floor plane using the input poses. Polygons whose overlap exceeds \(50\%\) of the smaller polygon area are grouped and merged by Shapely union. We also report a GT-pose variant that replaces predicted poses with ground-truth poses.

\paragraph{Topdown training data construction.}
\textit{RoomFormer}~\cite{yue2023roomformer} and \textit{PolyRoom}~\cite{liu2024polyroom} operate on topdown geometric inputs rather than panoramas, so we finetune both models on \texttt{ZInD-XV} training data constructed to mimic sparse multi-view observations. Starting from lifted floor-plans with fixed ceiling height and room connectivity preserved through open doors, we project per-panorama floor wall boundary points into the global frame using groundtruth poses and rasterize them into \(256\times256\) orthographic floor plane density maps. The corresponding room polygons from \texttt{ZInD-XV} are used as supervision. To match test-time conditions, we train with 2, 3, 5, and 8-view samples and add Gaussian noise with \(\sigma=0.001\) on boundary locations, \(\sigma=0.1\)\,m on panorama positions, and \(\sigma=0.05\)\,rad on panorama yaw.

\paragraph{RoomFormer.}
\textit{RoomFormer}~\cite{yue2023roomformer} is initialized from its \texttt{Structured3D} checkpoint with a \textit{ResNet-50} backbone, 800 vertex queries, and up to 20 room polygons per scene. We fine-tune \textit{RoomFormer} on the rendered \texttt{ZInD-XV} density maps and room polygon supervision described above, using AdamW with learning rate \(2\times 10^{-4}\) (\(2\times 10^{-5}\) for the backbone), weight decay \(10^{-4}\), batch size 10, and gradient clipping at max norm 0.1. Fine-tuning runs for approximately 70 epochs from the \texttt{Structured3D} initialization within a \textit{MultiStepLR} schedule whose learning-rate decay is configured at epoch 400 (not reached). At inference, we render the topdown density map from the evaluated input poses and predicted \textit{LGT-Net} floor wall boundaries, then apply the same vectorization pipeline: corners with foreground probability \(>0.5\) are retained, polygons with area \(<100\,\mathrm{px}^2\) are discarded, and surviving corners are mapped back to metric floor plane coordinates using a precomputed homography.

\paragraph{PolyRoom.}
\textit{PolyRoom}~\cite{liu2024polyroom} uses the same rendered input representation, but adds a \textit{Mask2Former}~\cite{cheng2021maskformer} room-instance segmentation stage whose predicted room contours initialize a \textit{RoomFormer} style polygon decoder instead of fixed learned queries. We train \textit{PolyRoom} in two stages: first fine-tuning a \textit{Mask2Former} with \textit{Swin-L} backbone initialized from \texttt{ImageNet-22K} pretraining, then fine-tuning the \textit{PolyRoom} decoder from its \texttt{Structured3D} checkpoint on the same rendered \texttt{ZInD-XV} data. Stage 1 trains \textit{Mask2Former} at \(256\times 256\) input resolution with AdamW (learning rate \(10^{-4}\), weight decay \(0.05\), batch size 2) for 400 epochs, with a backbone learning-rate multiplier of 0.1 and standard random-flip, random-resize, and random-crop augmentations. Stage 2 resumes from the \texttt{Structured3D}-pretrained \textit{PolyRoom} checkpoint, including its optimizer and scheduler state, and continues fine-tuning on \texttt{ZInD-XV} for the remainder of the original \textit{MultiStepLR} schedule (configured for 500 total epochs with a learning-rate decay at epoch 400) using AdamW (learning rate \(2\times 10^{-4}\), \(2\times 10^{-5}\) for the backbone), weight decay \(10^{-4}\), batch size 10, gradient clipping at max norm 0.1, 800 vertex queries, and up to 20 room polygons per scene. At inference, \textit{Mask2Former} instances with score \(>0.6\) are converted to contours using \textit{Douglas--Peucker} simplification \((\epsilon=5\,\mathrm{px})\) and used to seed the decoder. From the decoded polygons, we retain corners with foreground probability \(>0.1\) and interior angles within \(30^\circ\) of a right angle, merge nearby corners within 10 pixels, discard polygons with area \(<100\,\mathrm{px}^2\), and map the surviving polygons back to metric floor-plane coordinates.

\subsection{Layout Evaluation Metrics}
\label{sec:layout_metrics}
We evaluate wall and junction precision, recall, and \(F1\) using instance-level matching. A predicted wall is considered a candidate match if its orientation differs by at most \(10^\circ\) and its perpendicular distance is within \(2.0\) m of a ground-truth wall. Candidate wall pairs are then matched by \textit{Hungarian} assignment based on line overlap. Junction precision, recall, and \(F1\) are defined analogously: a junction is counted as matched when its two incident walls are both matched. We also report mean matched-wall localization error. We use these structure-aware metrics because \textit{MvFFN} predicts both closed polygons and open wall sequences, making region-based \(\mathrm{IoU}\) not well defined for all outputs.

\subsection{Downstream Refinement Protocol}
We evaluate planar bundle adjustment (\textit{BA})~\cite{zhou2023efficientplanarba} and \textit{BADGR}~\cite{li2025badgr} as downstream refinement methods applied to the \textit{MvFFN} layouts from Sec. 5.3 of the main paper. For each method, we report mean and median matched wall distances before and after refinement on the full evaluation set, and on the subset of samples whose mean matched wall distance improves; the latter also defines the improvement rate. Before refinement, partial room shapes are closed with a simple rule-based procedure to avoid self-intersections. This can introduce a small number of completed walls without image correspondences, which is especially challenging for \textit{BADGR} because it assumes the number of room walls is known. This explains why \textit{BADGR} is more sensitive to initialization in our setting.






\section{Failure Cases}
\label{sec:supp_failures}

\begin{figure}[H]
\begin{center}
    \includegraphics[width=0.9\linewidth]{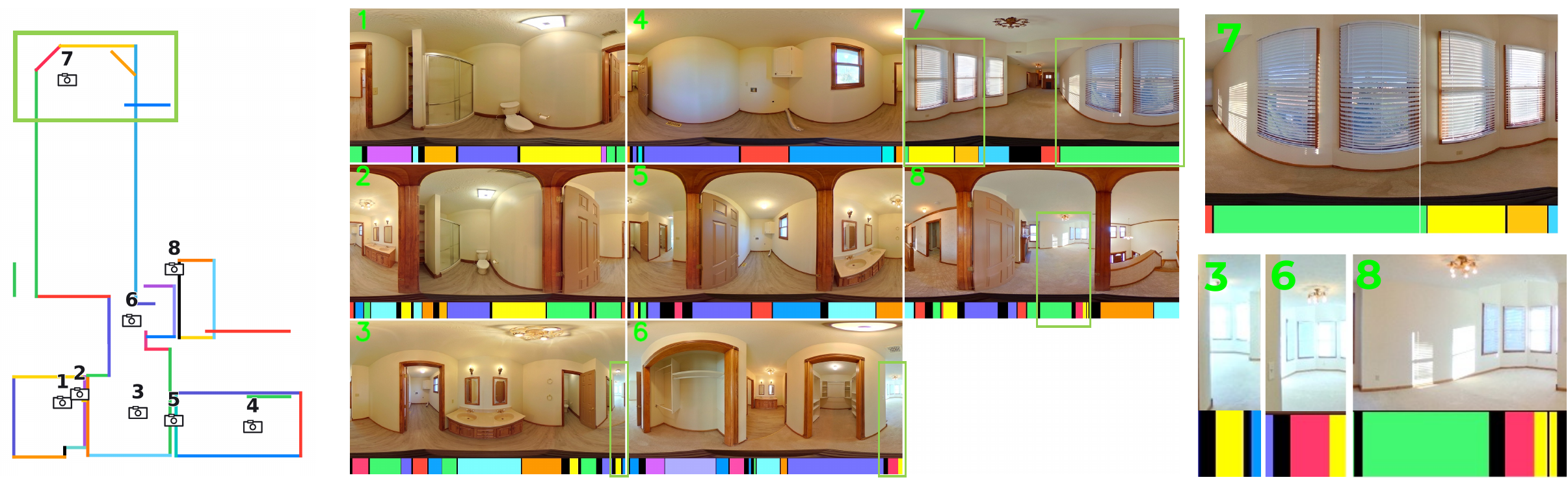}
    \caption{Failure case caused by incorrect \textit{MOIS} wall-instance prediction, where \textit{MvFFN} fails to distinguish different wall instances across views.}
    \label{fig:failure_1_conn}
\end{center}
\vspace{-16pt}
\end{figure}

A first failure mode of \textit{MvFFN} occurs when the predicted \textit{MOIS} field does not consistently separate wall instances across views, producing incorrect wall grouping. As shown in Fig.~\ref{fig:failure_1_conn}, these errors propagate through layout chaining and lead to both connectivity and geometric mistakes in the final layout. The failure likely stems from limited training coverage for non-Manhattan structures such as bay windows, where cross-view wall-instance prediction is more ambiguous. Although layout chaining suppresses weak wall hypotheses supported by only a few columns, noisy predictions from distant regions can still introduce incorrect connectivity. Inconsistent annotations for structures such as kitchen islands further worsen this problem under limited training data. More generally, because layout chaining is heuristic-based, it has limited ability to recover from noisy \textit{MOIS} predictions, highlighting the value of a learned post-processing module for robust global layout assembly.

\begin{figure}[H]
\begin{center}
    \includegraphics[width=0.75\linewidth]{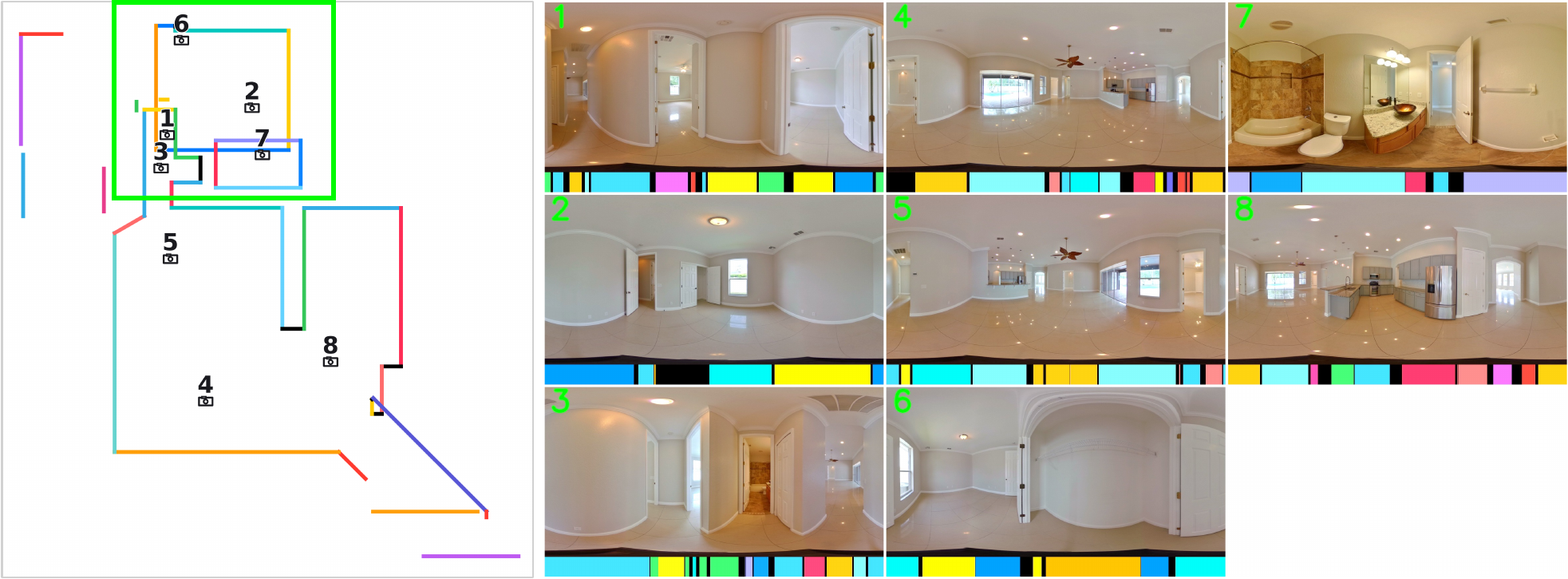}
    \caption{Failure case caused by geometric errors in the initial \textit{MvFFN} predictions, where inaccurate poses and point maps lead to distorted layout geometry.}
    \label{fig:failure_2_geom}
\end{center}
\vspace{-16pt}
\end{figure}

A second failure mode arises from geometric errors in the initial \textit{MvFFN} predictions, including inaccurate camera poses and noisy point maps, as shown in Fig.~\ref{fig:failure_2_geom}. Because layout chaining builds the global layout from these predicted geometric cues, such errors can propagate to the final reconstruction and produce misaligned walls or distorted room geometry. Although the wall-instance predictions of \textit{MvFFN} enable downstream bundle adjustment, the quality of the final refinement still depends on the accuracy of the initial geometry and correspondences.

\section{Limitations and Broader Impact}

\textbf{Limitations.}
\textit{MvFFN} assumes straightened equirectangular RGB panoramas from a single connected indoor pose graph, with sufficient overlap between neighboring views and known relative camera heights. Performance may degrade when these assumptions are violated, for example under extremely sparse captures, low co-visibility, severe occlusion, inaccurate camera-height metadata, strongly non-Manhattan or curved architecture, low-light or textureless images, or scenes outside the residential indoor domains evaluated here. Because structured multi-room training data remains limited, the current model uses two-view rotation and correspondence priors during training and inference, making performance sensitive to prior quality under domain shift. The method can return partial or open room layouts when visibility is insufficient, and downstream refinement can degrade when the initial layout or pixel-to-wall correspondences are noisy. The dense anchor-conditioned \textit{MOIS} inference and attention layers also scale with the number of views and image columns, so memory and runtime should be considered when applying the method to larger view sets.

\paragraph{Broader impact and responsible use.}
This work can support useful applications such as lower-cost indoor digital twins, AR/VR walkthroughs, assistive indoor navigation, and robotics mapping. At the same time, accurate reconstruction of private interior layouts can create privacy and security risks if used without consent, and erroneous reconstructions may be harmful in downstream navigation, robotics, or planning systems if treated as safety-critical. We therefore intend the method and released assets for research and consented indoor reconstruction, not for covert mapping, access-control, insurance, credit, law-enforcement, or other consequential decisions. Released annotations and code should follow the licenses and access restrictions of the underlying datasets, document intended use and limitations, and avoid distributing raw indoor imagery beyond what the original dataset terms permit.

\clearpage

\section{More Qualitative Results}
\label{sec:supp_quals}

\subsection{ZInD-XV Results}

\begin{figure}[H]
\begin{center}
    \includegraphics[width=\linewidth,height=0.88\textheight,keepaspectratio]{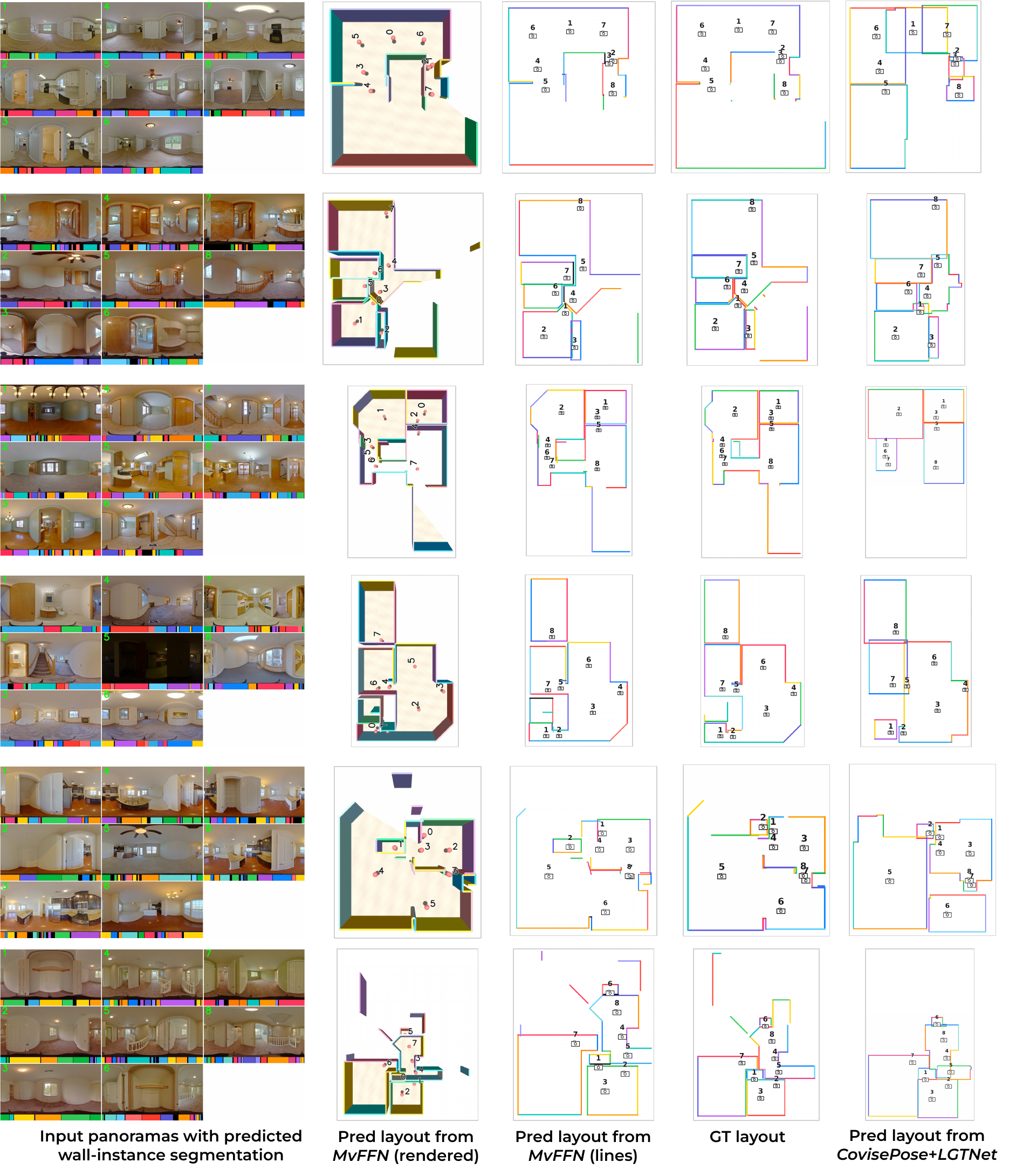}
    \caption{Qualitative comparisons on \texttt{ZInD-XV} test set.}
    \label{fig:quals_1}
\end{center}
\end{figure}

\begin{figure}[H]
\begin{center}
    \includegraphics[width=1.0\linewidth]{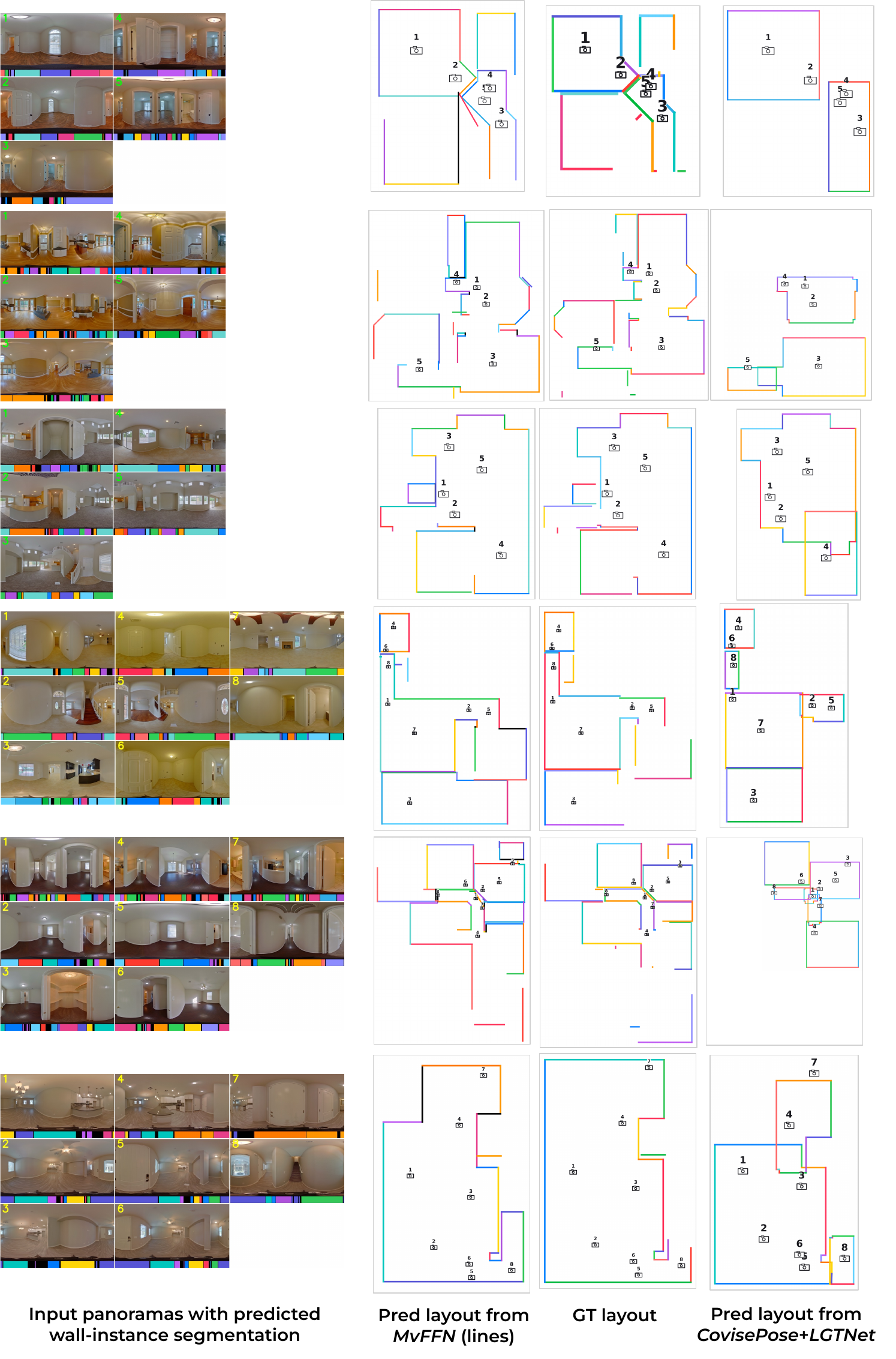}
    \caption{More Qualitative comparisons on \texttt{ZInD-XV} test set.} 
    \label{fig:quals_2}
    \vspace{-12pt}
\end{center}
\end{figure}

Figs.~\ref{fig:quals_1} and \ref{fig:quals_2} show additional qualitative examples on \texttt{ZInD-XV} from 5-view and 8-view samples. The first columns visualize the predicted multi-view wall instances with consistent colors across panoramas, followed by the predicted top-down layout, the ground-truth layout, and the strongest modular baseline, \textit{LGT-Net} with \textit{CovisPose}. Across both simple rooms and larger multi-room scenes, \textit{MvFFN} more reliably reconstructs complete room shapes, remains accurate under partial occlusion, and recovers distant wall instances and neighboring room structure visible through doorways. Consistent with the quantitative pose and layout results, the predicted camera poses and wall instances show fewer duplications and omissions than the baseline. In contrast, the baseline often predicts partial room shapes, splits large rooms into smaller fragments, and fails to reason jointly across views, leading to incorrect boundaries, missing wall structure, and redundant contours.

\subsection{HM3D-XV Results}

\begin{figure}[H]
\begin{center}
    \includegraphics[width=1.0\linewidth]{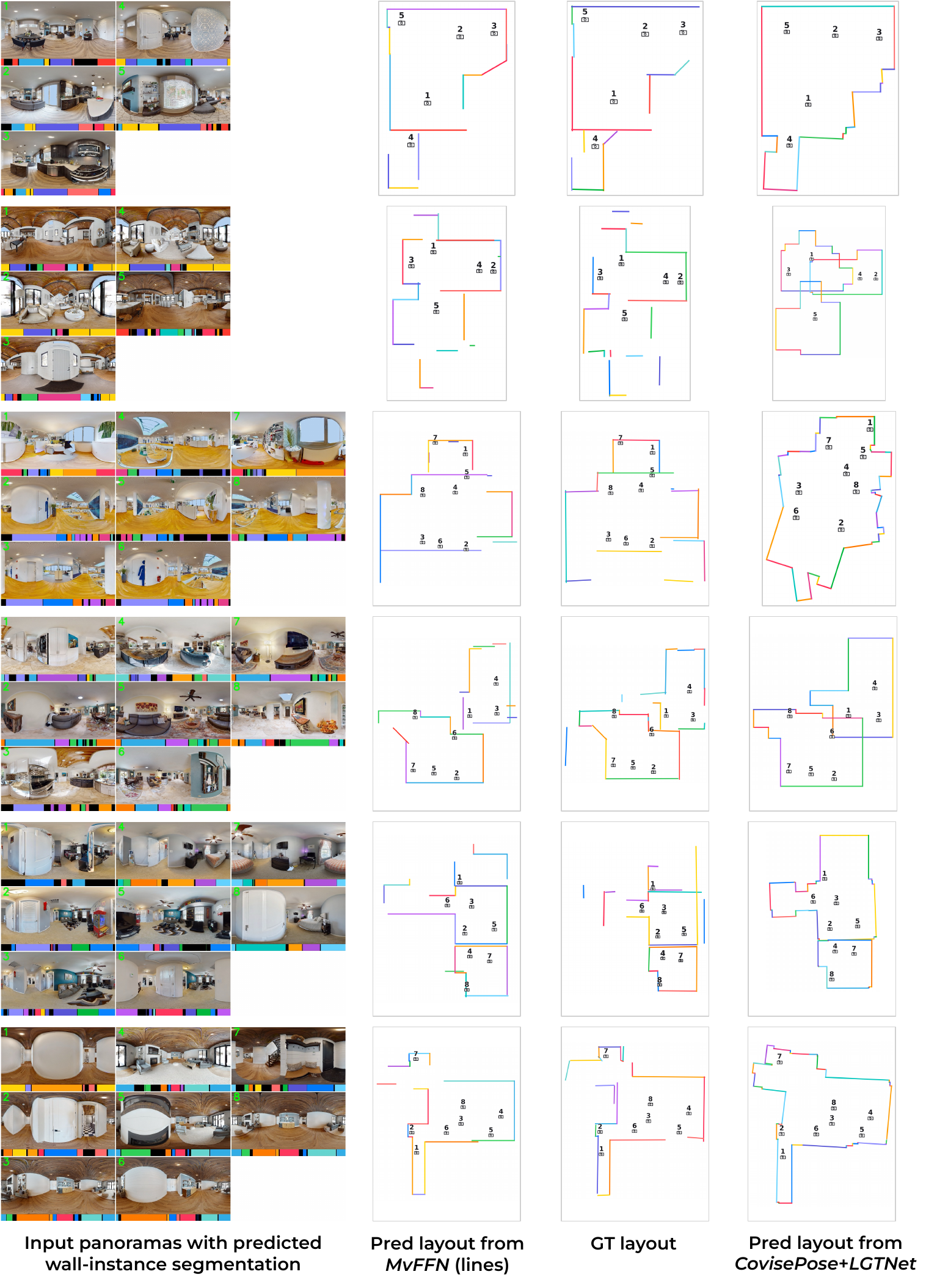}
    \caption{Qualitative comparisons on \texttt{HM3D-XV} test set.} 
    \label{fig:quals_hm3d}
    \vspace{-12pt}
\end{center}
\end{figure}

Figure~\ref{fig:quals_hm3d} shows \texttt{HM3D-XV} reconstruction examples under the furnished, mesh-rendered domain shift. As in \texttt{ZInD-XV}, \textit{MvFFN} can recover globally consistent wall instances and coherent room-level structure, reducing fragmented or duplicate walls relative to modular geometry-first baselines. Because \texttt{HM3D-XV} is smaller, noisier, and more sparsely annotated, these qualitative results illustrate representative successful cases rather than the full distribution of outcomes.


\section{Assets, Licenses, and Responsible Release}
\paragraph{Assets, licenses, and data release.}
This work uses existing datasets and research artifacts, including \texttt{ZInD}~\cite{cruz2021zillow}, \texttt{HM3D}/\texttt{HM3DSem}~\cite{yadav2023hm3dsem}, and the baseline or dependency code used in our comparisons. Table~\ref{tab:assets_licenses} lists the version or access date, citation, role, license or access terms, and release treatment for each asset, including pretrained weights and baseline implementations where applicable. \texttt{ZInD-XV} and \texttt{HM3D-XV} are derived annotation assets. Unless further restricted by the original dataset terms, we release our derived annotations, splits, metadata, and evaluation files under CC BY-NC 4.0. This license applies only to the annotations and metadata created in this work, not to the underlying raw panoramas, meshes, textures, renderings, or other source assets. We do not redistribute raw restricted source data; users must obtain \texttt{ZInD} and \texttt{HM3D}/\texttt{HM3DSem} from the original providers and comply with their respective licenses, terms of use, and access restrictions. We also provide conversion and evaluation scripts, together with a dataset card or README documenting label definitions, annotation and conversion pipelines, statistics, splits, file formats, known limitations, intended use, out-of-scope use, and maintenance policy.

\begin{table}[H]
\centering
\small
\setlength{\tabcolsep}{2pt}

\begin{tabular}{>{\raggedright\arraybackslash}p{1.5cm}>{\raggedright\arraybackslash}p{1.5cm}>{\raggedright\arraybackslash}p{2cm}>{\raggedright\arraybackslash}p{2cm}>{\raggedright\arraybackslash}p{6cm}}
\toprule
Asset & Role in this work & Version / access date & License or terms & Release treatment \\
\midrule


ZInD-XV annotations
& New derived annotation asset
& v1.0
& CC BY-NC 4.0, subject to original ZInD terms
& We release derived wall-instance labels, splits, metadata, conversion scripts, and evaluation files only. \\

HM3D-XV annotations
& New derived annotation asset
& v1.0
& CC BY-NC 4.0, subject to original HM3D and HM3DSem terms
& We release derived annotations, splits, metadata, conversion scripts, and evaluation files only. \\








\bottomrule
\end{tabular}
\caption{Existing and newly derived assets used in this work. We release only assets created by this work under the stated terms and do not redistribute raw restricted source data.}
\label{tab:assets_licenses}
\end{table}

\paragraph{Dataset card summary.}
\texttt{ZInD-XV} contains derived multi-view floor-plan annotations, globally indexed wall instances, room-wise wall sequences, and per-panorama column-wise floor-boundary and wall-instance labels. \texttt{HM3D-XV} contains analogous derived annotations for residential \texttt{HM3DSem} scenes. The released files include train/validation/test splits, per-floor metadata, per-panorama labels, wall-instance identifiers, room connectivity, and evaluation tuples. The README specifies the file format, coordinate convention, label definitions, preprocessing steps, conversion scripts, and evaluation protocol. The intended use is research on indoor layout reconstruction, camera pose estimation, and structured multi-view scene understanding. Out-of-scope uses include covert mapping of private spaces, security-sensitive reconstruction without consent, and consequential decision-making based on predicted layouts.

\paragraph{Responsible release.}
Indoor floor-plans can reveal sensitive information about private spaces. To reduce privacy, security, and license risks, we do not redistribute raw restricted panoramas, meshes, textures, renderings, or floor-plan source files. The released package contains only compatible derived annotations, metadata, splits, and scripts. Users are required to obtain the original datasets from their providers and comply with the corresponding terms. The release documentation states intended uses, out-of-scope uses, limitations, and access restrictions. We will release all code, trained checkpoints, and utilities upon acceptance.

\paragraph{Annotation provenance.}
Manual annotation and cleanup for \texttt{ZInD-XV} and \texttt{HM3D-XV} were performed by the authors/research team using custom annotation and quality-control tools. No crowdworkers, contractors, or human-subject study participants were recruited. The released documentation describes the annotation interface, label definitions, quality-control procedure, and cleanup criteria.
\section{Reproducibility and Compute}
\label{app:repro_compute}

\paragraph{Compute resources.}
All \textit{MvFFN} experiments were run on 8 or 16 A100 GPUs with 40GB memory per GPU, using 6~12 cpus per gpu. Note that the experiments can also be conducted on lower-tier gpus. Stage-1 geometry pre-training used approximately 2 days for 80k iterations, and stage-2 layout semantic training used approximately 2.5 days for 80k iterations. We used a batch size of [10,7,4,2] for 2, 3, 5, 8 views with 6~12 dataloader workers per process. 



\ifdefined\neuripscameraready
  \clearpage
  \section*{NeurIPS Paper Checklist}

\begin{enumerate}

\item {\bf Claims}
    \item[] Question: Do the main claims made in the abstract and introduction accurately reflect the paper's contributions and scope?
    \item[] Answer: \answerYes{}
    \item[] Justification: The abstract and introduction state the paper's main contributions: the \textit{MOIS} formulation, the \textit{MvFFN} model for predicting local cross-view wall structures, and the \texttt{ZInD-XV} dataset. They also describe how local predictions are assembled into global layouts and how the resulting wall correspondences are evaluated for downstream refinement. The experiments support these claims on sparse, unposed indoor panorama layout reconstruction. The scope is limited to the stated indoor 360$^\circ$ panorama setting, including straightened panoramas, connected pose graphs, sufficient co-visibility, and known relative camera heights, as discussed in the method and the supplementary limitations.

\item {\bf Limitations}
    \item[] Question: Does the paper discuss the limitations of the work performed by the authors?
    \item[] Answer: \answerYes{}
    \item[] Justification: The supplementary limitations section discusses the main assumptions and failure modes, including straightened equirectangular panoramas, a connected indoor pose graph, sufficient overlap, known relative camera heights, low co-visibility, occlusion, non-Manhattan or curved architecture, low-light or textureless images, partial/open layouts, refinement sensitivity, and scaling with the number of views and image columns.

\item {\bf Theory assumptions and proofs}
    \item[] Question: For each theoretical result, does the paper provide the full set of assumptions and a complete (and correct) proof?
    \item[] Answer: \answerNA{}
    \item[] Justification: The paper does not present formal theoretical results, theorems, or proofs; the mathematical notation defines the representation, architecture, losses, and optimization procedures.

\item {\bf Experimental result reproducibility}
    \item[] Question: Does the paper fully disclose all the information needed to reproduce the main experimental results of the paper to the extent that it affects the main claims and/or conclusions of the paper (regardless of whether the code and data are provided or not)?
    \item[] Answer: \answerYes{}
    \item[] Justification: The method and the supplementary material describe the architecture, MOIS targets, two-stage training, Layout Chaining, refinement, dataset construction, tuple construction, metrics, baselines, losses, optimizer settings, sampling schedules, batch sizes, and compute setup needed to reproduce or verify the main empirical claims.

\item {\bf Open access to data and code}
    \item[] Question: Does the paper provide open access to the data and code, with sufficient instructions to faithfully reproduce the main experimental results, as described in supplemental material?
    \item[] Answer: \answerNo{}
    \item[] Justification: We do not provide open-access code with the NeurIPS submission. The paper and the supplementary material provide the method, training, evaluation, dataset-construction, and baseline details needed to understand and verify the main results. We commit to releasing code upon acceptance. Raw \texttt{ZInD} and \texttt{HM3D}/\texttt{HM3DSem} source data are not redistributed due to the original dataset terms. We also plan to release derived \texttt{ZInD-XV} and \texttt{HM3D-XV} annotations, metadata, splits, conversion scripts, evaluation files, and potentially the annotation GUI under terms compatible with the original datasets.

\item {\bf Experimental setting/details}
    \item[] Question: Does the paper specify all the training and test details (e.g., data splits, hyperparameters, how they were chosen, type of optimizer) necessary to understand the results?
    \item[] Answer: \answerYes{}
    \item[] Justification: The experimental sections specify the evaluated datasets, view-count settings, baselines, metrics, tuple construction, and main ablations. The supplementary material provides training losses, optimizer settings, sampling schedules, dataset preparation, Layout Chaining details, baseline implementations, evaluation metrics, batch sizes, and compute resources.

\item {\bf Experiment statistical significance}
    \item[] Question: Does the paper report error bars suitably and correctly defined or other appropriate information about the statistical significance of the experiments?
    \item[] Answer: \answerYes{}
    \item[] Justification: The paper reports distributional summaries for the pose experiments, including mean, standard deviation, and median over evaluated panoramas, and reports layout and refinement results as aggregate metrics over fixed test tuple sets. These summaries characterize variability across evaluated examples, although the paper does not report multi-seed training variability or formal hypothesis tests.

\item {\bf Experiments compute resources}
    \item[] Question: For each experiment, does the paper provide sufficient information on the computer resources (type of compute workers, memory, time of execution) needed to reproduce the experiments?
    \item[] Answer: \answerYes{}
    \item[] Justification: The supplementary material reports the compute resources used for the main MvFFN experiments, including GPU type, GPU memory, number of GPUs, CPU/dataloader resources, training time for both stages, iteration counts, and view-dependent batch sizes. These details provide the main hardware and runtime requirements needed to reproduce the reported training setup.
    
\item {\bf Code of ethics}
    \item[] Question: Does the research conducted in the paper conform, in every respect, with the NeurIPS Code of Ethics \url{https://neurips.cc/public/EthicsGuidelines}?
    \item[] Answer: \answerYes{}
    \item[] Justification: The research conforms to the NeurIPS Code of Ethics to the best of the authors' knowledge: it uses existing indoor-scene datasets and author-created derived annotations, does not involve deception or weaponization, documents privacy and security risks, respects original dataset access restrictions, and uses a responsible-release strategy for derived assets.

\item {\bf Broader impacts}
    \item[] Question: Does the paper discuss both potential positive societal impacts and negative societal impacts of the work performed?
    \item[] Answer: \answerYes{}
    \item[] Justification: The supplementary broader-impact discussion covers positive impacts such as indoor digital twins, AR/VR walkthroughs, assistive indoor navigation, and robotics mapping, as well as negative impacts such as privacy and security risks from interior reconstruction, unauthorized mapping, and errors in downstream navigation, robotics, or planning systems.
    
\item {\bf Safeguards}
    \item[] Question: Does the paper describe safeguards that have been put in place for responsible release of data or models that have a high risk for misuse (e.g., pre-trained language models, image generators, or scraped datasets)?
    \item[] Answer: \answerYes{}
    \item[] Justification: Although the model is not a high-risk generative model, indoor layout reconstruction can raise privacy and security concerns. The supplementary material describes safeguards including derived-annotation-only distribution where required, no redistribution of raw restricted panoramas, meshes, textures, renderings, or floor plan source files, user compliance with original dataset terms, intended-use documentation, out-of-scope-use documentation, and release limitations.

\item {\bf Licenses for existing assets}
    \item[] Question: Are the creators or original owners of assets (e.g., code, data, models), used in the paper, properly credited and are the license and terms of use explicitly mentioned and properly respected?
    \item[] Answer: \answerYes{}
    \item[] Justification: The supplementary assets and licenses section documents the use of existing datasets and research artifacts, including ZInD, HM3D/HM3DSem, and baseline or dependency code, and describes the license or access restrictions that govern the released derived assets. The derived ZInD-XV and HM3D-XV annotations are released under CC BY-NC 4.0 only where compatible with the original dataset terms, and raw restricted source data are not redistributed.

\item {\bf New assets}
    \item[] Question: Are new assets introduced in the paper well documented and is the documentation provided alongside the assets?
    \item[] Answer: \answerYes{}
    \item[] Justification: The paper introduces ZInD-XV and HM3D-XV derived annotation assets, and the supplementary material documents their construction, label definitions, released files, statistics, splits, file formats, access restrictions, intended use, out-of-scope use, known limitations, license, responsible-release policy, and annotation provenance.
    \begin{itemize}
        \item The answer \answerNA{} means that the paper does not release new assets.
        \item Researchers should communicate the details of the dataset\slash code\slash model as part of their submissions via structured templates. This includes details about training, license, limitations, etc. 
        \item The paper should discuss whether and how consent was obtained from people whose asset is used.
        \item At submission time, remember to anonymize your assets (if applicable). You can either create an anonymized URL or include an anonymized zip file.
    \end{itemize}

\item {\bf Crowdsourcing and research with human subjects}
    \item[] Question: For crowdsourcing experiments and research with human subjects, does the paper include the full text of instructions given to participants and screenshots, if applicable, as well as details about compensation (if any)? 
    \item[] Answer: \answerNA{}
    \item[] Justification: The paper does not report crowdsourcing experiments or human-subject studies. Manual annotation and cleanup for ZInD-XV and HM3D-XV were performed by the authors/research team using custom annotation and quality-control tools; no crowdworkers, contractors, or human-subject study participants were recruited.

\item {\bf Institutional review board (IRB) approvals or equivalent for research with human subjects}
    \item[] Question: Does the paper describe potential risks incurred by study participants, whether such risks were disclosed to the subjects, and whether Institutional Review Board (IRB) approvals (or an equivalent approval/review based on the requirements of your country or institution) were obtained?
    \item[] Answer: \answerNA{}
    \item[] Justification: The study does not involve direct interaction with human participants or collection of new human-subject data; it uses existing indoor-scene datasets and author-created derived layout annotations.

\item {\bf Declaration of LLM usage}
    \item[] Question: Does the paper describe the usage of LLMs if it is an important, original, or non-standard component of the core methods in this research? Note that if the LLM is used only for writing, editing, or formatting purposes and does \emph{not} impact the core methodology, scientific rigor, or originality of the research, declaration is not required.
    \item[] Answer: \answerNA{}
    \item[] Justification: LLMs are not an important, original, or non-standard component of MvFFN, dataset construction, or the experimental methodology; any use for writing, editing, or formatting does not affect the core scientific contribution.

\end{enumerate}

\fi

\end{document}